\documentclass[lettersize,journal]{IEEEtran}

\usepackage{amsmath,amsfonts}
\usepackage{algorithmic}
\usepackage{array}
\usepackage{textcomp}
\usepackage{stfloats}
\usepackage{url}
\usepackage{verbatim}
\usepackage{graphicx}
\def\BibTeX{{\rm B\kern-.05em{\sc i\kern-.025em b}\kern-.08em
    T\kern-.1667em\lower.7ex\hbox{E}\kern-.125emX}}
\usepackage{balance}

\usepackage{hyperref}
\usepackage{orcidlink}
\usepackage{float}
\usepackage{caption}
\usepackage{subcaption}
\usepackage{booktabs}
\usepackage{tabulary}
\usepackage{makecell}
\usepackage{enumerate}
\usepackage{gensymb}

\newcommand{\CoDA}{\emph{CoDA}}
\newcommand{\CoDAGraph}{\emph{CoDA.Graph}}

\begin{document}

% \title{CorA -- A Coral Analyzer:\\Interactive Segmentation and Morphological Exploration of Stony Corals}
\title{CoDA: Interactive Segmentation and Morphological Analysis of Dendroid Structures Exemplified on Stony Cold-Water Corals}

\author{
    Kira Schmitt, Jürgen Titschack, and Daniel Baum
%    \thanks{TODO: Insert contributors we want to thank or some other subtext.}
}
% \markboth{Journal of \LaTeX\ Class Files,~Vol.~18, No.~9, September~2020}{How to Use the IEEEtran \LaTeX \ Templates}
\markboth{arXiv preprint}{arXiv preprint}
\maketitle

%% -- Teaser
\begin{figure}[ht!]
    \centering
    \includegraphics[width=\columnwidth]{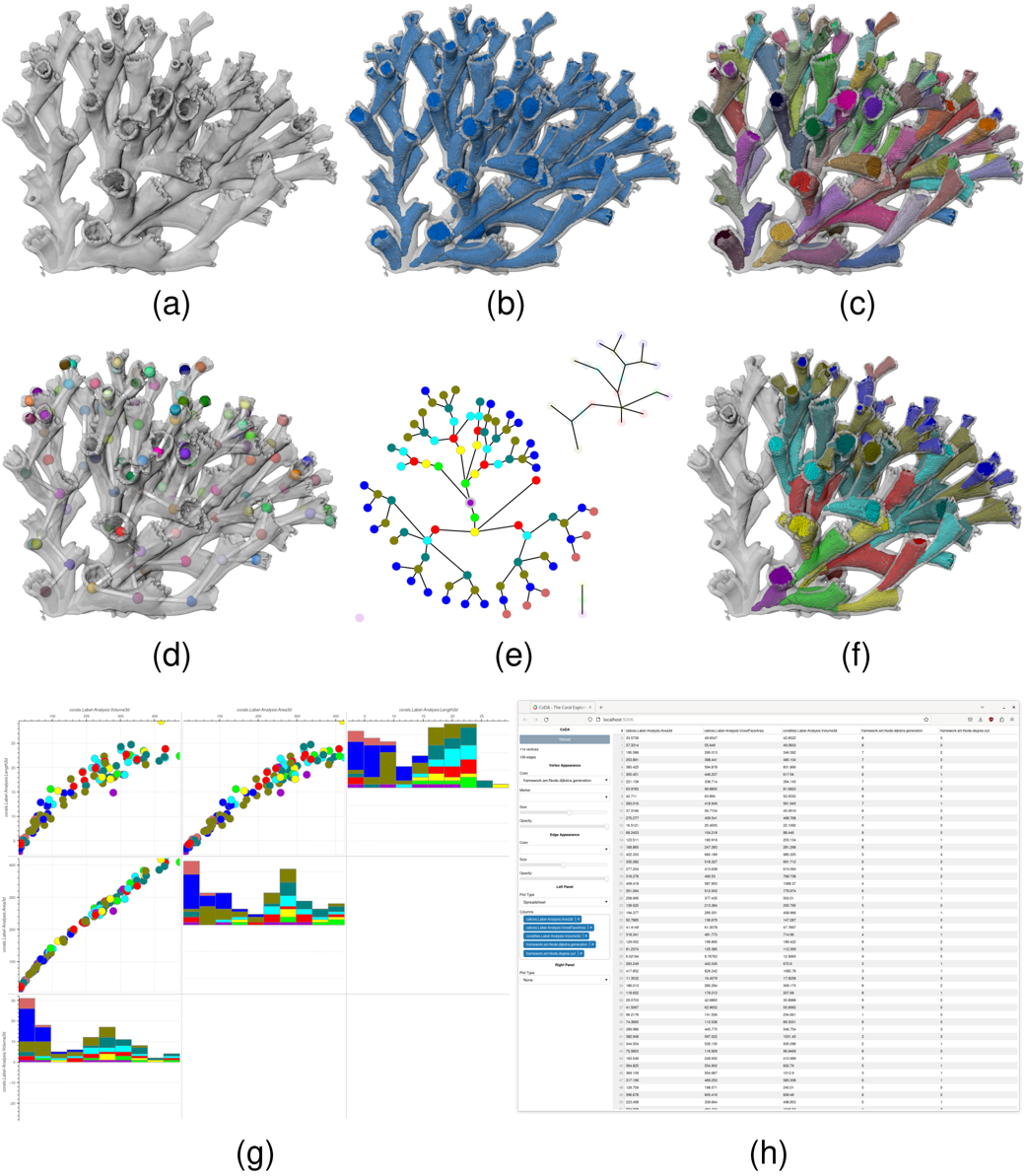}
    \caption{
        \CoDA\ enables the analysis and visual exploration of the ontogenetic morphological development of dendroid cold-water corals.
        \emph{(a)}%
            ~The analysis starts with a computed-tomography (CT) image of the cold-water coral (CWC).
        \emph{(b, c)}%
            ~The polyp-cavity (calyx) segmentation~\cite{2021:schmitt,2016:Baum,2018:Titschack} is the basis for the subsequent calyx/corallite instance segmentation.
        \emph{(d)}%
            ~Within the \CoDA\ framework, a skeleton tree, describing the mother-daughter corallite relationship within a colony, is fitted to the corals.
        \emph{(e, f)}%
            ~The skeleton tree is visualized as graph layout in \CoDAGraph, our 2D visualization suite that augments \CoDA\ with abstract, linked views, allowing the user to explore the colony and to focus on specific subsets of the tree during the analysis, e.g., on the descendants or ancestors of a calyx/corallite.
        \emph{(g, h)}%
            ~Traditional plots and views, e.g., scatter plot matrices (SPLOMs) and spreadsheets available in \CoDAGraph, complement \CoDA's exploration and analysis capabilities.
    }
    \label{fig:teaser}
\end{figure}

%% -- Abstract
\begin{abstract}
Dendroid stony corals build highly complex colonies that develop from a single coral polyp sitting in a cup-like skeleton, called corallite, by asexual reproduction, resulting in a tree-like branching pattern of its skeleton.
Despite their beauty and ecological importance as reef builders in tropical shallow-water reefs as well as in cold-water coral mounds in the deep ocean, systematic studies investigating the ontogenetic morphological development of such coral colonies are largely missing.
One reason for this is the sheer number of corallites – up to several thousands in a single coral colony.
Another limiting factor, especially for the analysis of dendroid cold-water corals, is the existence of many secondary joints in the ideally tree-like structure that make a reconstruction of the skeleton tree extremely tedious. 

Herein, we present \CoDA, the \emph{Co}ral \emph{D}endroid structure \emph{A}nalyzer, a visual analytics suite that allows for the first time to investigate the ontogenetic morphological development of complex dendroid coral colonies, exemplified on three important framework-forming dendroid cold-water corals: \textit{Lophelia pertusa} (Linnaeus, 1758), \textit{Madrepora oculata} (Linnaeus, 1758), and \textit{Goniocorella dumosa} (Alcock, 1902). 
Input to \CoDA\ is an initial instance segmentation of the coral polyp cavities (calices), from which it estimates the skeleton tree of the colony and extracts classical morphological measurements and advanced shape features of the individual corallites. 
\CoDA\ also works as a proofreading and error correction tool by helping to identify wrong parts in the skeleton tree and providing tools to quickly correct these errors.
The final skeleton tree enables the derivation of additional information about the calices/corallite instances that otherwise could not be obtained, including their ontogenetic generation and branching patterns -- the basis of a fully quantitative statistical analysis of the coral colony morphology.
Part of \CoDA\ is \CoDAGraph, a feature-rich link-and-brush user interface for visualizing the extracted features and 2D graph layouts of the skeleton tree, enabling the real-time exploration of complex coral colonies and their building blocks, the individual corallites and branches.

In the future, we expect \CoDA\ to greatly facilitate the analysis of large stony corals of different species and morphotypes, as well as other dendroid structures, enabling new insights into the influence of genetic and environmental factors on their ontogenetic morphological development.
\end{abstract}


%% -- Introduction
\section{Introduction}
\label{sec:introduction}

\begin{figure*}
    \centering
    \begin{subfigure}[t]{0.38\textwidth}
        \includegraphics[width=\textwidth]{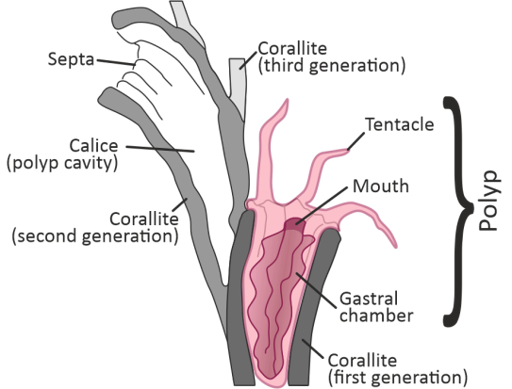}
        \caption{}
        \label{subfig:cold water coral scheme}
    \end{subfigure}
    \hfill
    \begin{subfigure}[t]{0.3\textwidth}
        \includegraphics[width=\textwidth]{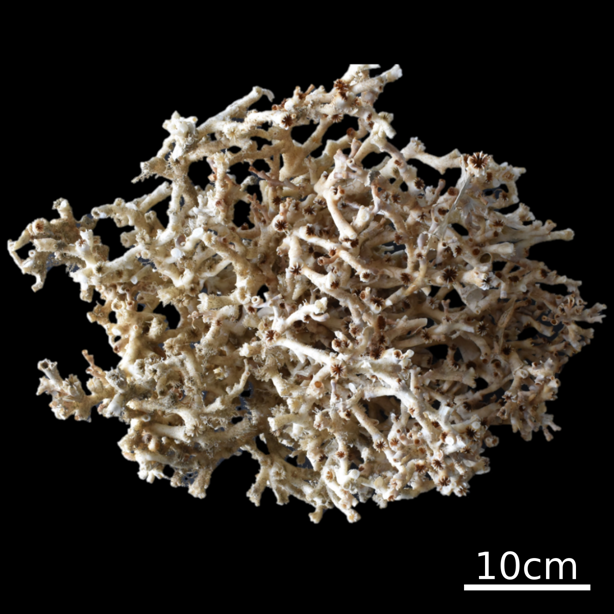}
        \caption{}
        \label{subfig:photo SaM-ID43148}
    \end{subfigure}
    \hfill
    \begin{subfigure}[t]{0.3\textwidth}
        \includegraphics[width=\textwidth]{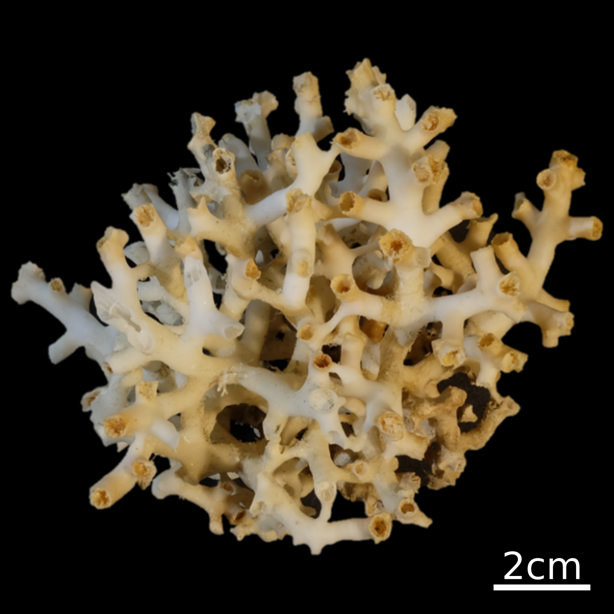}
        \caption{}
        \label{subfig:photo NIWA-148046}
    \end{subfigure}
    % TODO: @Jürgen, who owns the copyright/license to the pictures? If we don't state anything here, it could end up belonging to IEEE.
    \caption{
        \emph{(a)}
            The schematic depicts the typical parts of a cold-water coral colony (CC~BY~4.0~Deed~\cite{2021:schmitt}).
            The terms \emph{calyx} and \emph{calice} (in the schematic) can be used interchangeably and refer to the same part.
            We prefer \emph{calyx} in the singular, while the plural is \emph{calices}.
        \emph{(b)}
            The photo shows the \emph{SaM-ID43148 (III)} cold-water coral colony specimen.
            It is particularly large compared to other datasets and contains several colony fragments. 
            Its dense structure makes the analysis and visual exploration particularly challenging without adequate tools.
        \emph{(c)}
            The photo shows the \emph{NIWA-148046 (V)} cold-water coral colony specimen.
            Its morphology differs from the other corals presented in this paper as the cavities resemble more a cylinder than a cone.
    }
    \label{fig:background information}
\end{figure*}

\IEEEPARstart{C}{omplex} dendroidal structures are widespread in nature and commonly formed by various organisms comprising plants, e.g., trees, grass, including their roots, coralline algae (e.g.,\cite{2021:roth,2023:gonin,2022:granse,2007:hulzen, 2022:lines, 2024:cabrito}), and (marine) animals, e.g., corals, including the herein investigated cold-water corals (CWC), or bryozoans (\cite{2023:gonin,2017:orejas,2023:achilleos}).
Often, these organisms are key players in ecosystems, such as trees in forests or corals in coral reefs, by forming multiple ecological niches beneficial for other organisms, so that these ecosystems represent biodiversity hotspots and provide important ecosystem services (e.g., \cite{2021:taye,2014:armstrong,2023:cordes,2024:giglio}). 

Coral colonies consist of multiple coral polyps that originate from asexual budding.
Each polyp produces a skeleton called corallite with a central cavity, called calyx, in which it lives~(\autoref{subfig:cold water coral scheme}). 
Coral colonies exhibit various shapes (e.g., encrusting, hemispherical, tabular, corymbose and branching; \cite{2020:cresswell}) but colonial stony (scleactinian) cold-water corals are predominantly branching (dendroidal; \cite{2009:roberts}). 
Some dendroid cold-water corals commonly show secondary joints, regions in the colony where two different coral branches have grown together, and skeletal strengthening as well as further associated secondary joints related to the presence of worms of the genus \textit{Eunice} \cite{2013:mueller}. 
Both represent important features in the analysis of cold-water coral colonies as they influence significantly the morphology of the colonies, which is the result of the interaction of the organism with its ambient environment. 
However, an in-depth understanding of this interaction is often hindered by the limited methodologies to characterize three-dimensional morphologies. 
Often, morphologies are only qualitatively described (e.g., for CWC: \cite{1997:freiwald,2009:brooke,2020:chapron}) or characterized by some basic measurements (e.g., for CWC:~\cite{2021:Sanna,2023:sanna}).

This work presents a new visual analytics suite called \CoDA\ that allows a detailed three-dimensional quantitative assessment of dendroidal structures. 
A major component of \CoDA\ is \CoDAGraph, a browser-based application for showing 2D plots like scatter plots, SPLOMs, maps and graph layouts.
\CoDAGraph\ is not only suited for the analysis of CWCs but any kind of dendroid structures.
We tested \CoDA\ on colony fragments of the reef-forming CWCs \textit{Lophelia pertusa}, \textit{Madrepora oculata} and \textit{Goniocorella dumosa}.
An overview of the analysis workflow using \CoDA\ is shown in Figure~\ref{fig:teaser}.
Note that in this work, we visualize all binary and instance segmentations as voxelized data for two reasons:
(1)~It gives the reader a better feeling for the resolution of the data.
(2)~It is the one used in \CoDA\ as it is more interactive than using a surface representation that needs recomputation with each change.

\subsection{Related work}
\label{subsec:related work}

Skeleton graphs and center line trees~\cite{2000:sato,2006:fouard} are well established tools in the image processing community and provide a good first impression of the growth of a coral colony.
However, they are purely based on topological measures and do not take into account the actual, biological relationship.
They are thus not suited for a proper ontogenetic analysis.
The work on \CoDA\ is motivated by the desire to obtain a \emph{directed} skeleton graph in which the edges coincide with the mother-daughter relationships between adjacent corallites and in which each vertex corresponds to a corallite.
For similar reasons, contact graphs~\cite{2022:pandey}, as they occur in the analysis of particle and fibre systems, are not suited for the ontogenetic analysis.

Our 2D analytics platform \CoDAGraph\ is inspired by \emph{dExplorer}~\cite{2023:choi}, an interactive platform for spine analysis.
It allows the extraction of many morphological features and their visualization in a single application.
However, with regard to the analysis of CWC colonies, we are interested in different features, both for individual corallites as well as morphological features of the entire coral colony.
Most importantly, we are interested in the visualization and exploration of the relationships within a colony.

The visualization and modification of the graph structures are based on the work of Dercksen et~al.\ \cite{2013:dercksen}, in which they deal with the tracing and visualization of 3D dendrite and axon morphologies. 
We apply many of the presented ideas, especially the manual tools for editing the spatial graphs, in our analysis pipeline, extend and adjust them to our use case.
Dercksen et~al.\ construct the edges by tracing the dendrites in the image.
For coral colonies, this is not possible, as there is no unique path visible in the image that connects mother and daughter corallites (\autoref{subfig:cold water coral scheme}).
Additionally, the edges in the dendrite graph are again undirected whereas the orientation of the edges carry important information in our application.

\subsection{Data}
\label{subsec:data}

This paper explores five specimens of, in total, three different species of cold-water corals with different morphologies.
All specimens vary greatly in size, so that we not only test the developed methods for small samples, but also for large ones with almost 2000 corallites.
We consider the following specimens:
\begin{enumerate}[(I)]
    \item
        % Jürgens mail:
        %   63°36.46′N and 9°22.76′E and 157 water depth)
        %   Leksa reef
        %   63°36.46′N 
        %   9°22.76′E 
        %   157 m water depth
        sample \emph{A2W} of the species \textit{Lophelia pertusa} with elongated corallites and an open, bushy colony morphology, collected from Leksa reef, Norway ($63.61\degree\,\text{N}$ $9.38\degree\,\text{E}$ and 157m water depth);
    \item 
        % ambient curvature paper: 
        %   (64.1110°N, 8.1187°E, 303m)
        % Jürgens mail:
        %   (64.1110°N, 8.1187°E, -303m)
        %   Sula reef
        %   Submersible JAGO (JAGO)
        sample \emph{C1W} of the species \textit{Lophelia pertusa} with elongated corallites and an open, bushy colony morphology, collected from Sula reef, Norway ($64.11\degree\,\text{N}$ $8.12\degree\,\text{E}$ in 303m depth);
    \item 
        % ambient curvature paper: 
        %   (41°43.51' N, 17°02.78' E, 710m)
        %   ROV MARUM-Quest
        %   Gondola Slide, southeast of the Gargano Promontory in the Adriatic Sea
        %   stations M70/1-752, GeoB11207
        %
        % label in picture:
        %   VH-24-95
        %   stations 166+167        (!!!! MISMATCH)
        %
        % Jürgens mail:
        %   GS: M70/1-752 (D 111)
        %   Latitude start: 41.725170
        %   Longitude Start: 17.046330
        %   Latitude End: 41.719670
        %   Longitude End: 17.060830
        %   Date/Time: 2006-10-15T07:11:00
        %   Elevation Start: -674.0 m 
        %   Elevation End: -710.0 m 
        %   Location: Mediterranean Sea 
        %   Basis: Meteor (1986)
        %   Method: ROV
        %   Comment: GeoB11207
        sample \emph{SaM-ID43148} of the species \textit{Lophelia pertusa} with slender corallites and a columnar colony morphotype, collected near the Gondola Slide, southeast of the Gargano Promontory in the Adriatic Sea, (stations 167/168; GeoB11207; $41.72\degree\,\text{N}$ $17.05\degree\,\text{E}$, approximately in 700m water depth; \cite{1995:freiwald}), shown in \autoref{subfig:photo SaM-ID43148};
    \item 
        % Jürgens mail:
        %   ROV
        %   2008-03-01 11:30
        %   LAT 34.98833
        %   LONG -7.07200
        %   DEPTH -714
        sample \emph{GeoB12747-1} of the species \textit{Madrepora oculata} with its typical zigzag corallite arrangement, forming a fragile, predominantly fan-shaped colony morphology, collected from Meknes mud vulcano, Gulf of Cadiz during Pelagia cruise 64PE284 \cite{2008:hebbeln}
        ($34.99\degree\,\text{N}$ $7.07\degree\,\text{W}$ in 714m depth);
    \item 
        % label on picture: 
        %   (43.369°S 179.452°E, 394m-402m)
        and sample \emph{NIWA-148046} of the species \textit{Goniocorella dumosa} with its tube-like corallites and bushy colony morphology, collected from Chatham Rise, New Zealand ($43.37\degree\,\text{S}$ $179.45\degree\,\text{E}$ in approximately 400m water depth), shown in \autoref{subfig:photo NIWA-148046}.
\end{enumerate}
The CT scans were performed at Klinikum Bremen-Mitte with the same protocol as described in Schmitt~et~al.~(\cite{2021:schmitt}, specimens (I), (II) and (IV) with the Toshiba Aquilion 64 device, and specimens (III) and (V) with the Philips Brilliance iCT Elite 256 device).
The qualitative classification scheme of Sanna~et~al.~\cite{2021:Sanna} was used to determine colony morphotypes.
More information regarding meta data and results are provided in \autoref{tab:num edges num vertices benchmark}.

\subsection{Contributions}
\label{subsec:contributions}

% TODO: @Daniel and @Jürgen, please take a look this, I think I've gone a little overboard with the listing and I'm not sure how to make it more consice right now. 
Our contributions are as follows.
\begin{itemize}
    \item 
        A domain-specific framework for the semi-automatic ontogenetic and morphological analysis of cold-water coral colonies, potentially suitable for other dendroidal structures.
    \item 
        A linked dendrogram and state-of-the-art graph layouts for the coral colonies that provide an abstract, unobtrusive view of the colonies' growth patterns; it also provides multiple selection tools for exploring the ancestry of individual corals and their ontogenetic development.
    \item 
        A heuristic for orienting cold-water coral calices/corallites, and other cone-shaped structures, that approximates their centerlines by parabolas; a further heuristic that uses the orientation information to estimate the skeleton tree of coral colonies.
    \item 
        A proofreading mechanism for the instance segmentation of the calices/corallites and their adjacency, which avoids occlusions and guides the user in a coherent way through all instances, even in very large colonies.
    \item 
        A tool set for jointly correcting over-segmentations, under-segmentations and wrong adjacencies in the instance segmentation and the attached graph, in a systematic manner without having to start all over again.
    \item 
        A simple mechanism for linking and brushing between different applications using a simple \emph{.csv} file-based interface, that allows the seamless integration of \CoDAGraph\ into other visualization tools.
\end{itemize}

The remainder of the paper is structured as follows:
In \autoref{sec:segmentation workflow}, we give an overview of the segmentation workflow, the fitting of the colony skeleton graph and our tools for jointly refining the calyx instance segmentation and the skeleton graph.
In \autoref{sec:implementation}, we present the visual analysis tool \CoDAGraph, describe its architecture, its features and its integration into other visualization platforms.
In \autoref{sec:case studies}, we apply the segmentation workflow to five different coral specimens of three cold-water coral taxa, explore the results with \CoDA, describe the overall workflow and experience while processing the samples.

%% -- Segmentation workflow
\section{Segmentation Workflow}
\label{sec:segmentation workflow}

Input to our segmentation pipeline is a 3D image resulting from scanning individual specimens containing one or several coral colony fragments using computed tomography.
An iso-surface rendering of such a 3D image is shown in \hyperref[fig:teaser]{Figure~1a}.

\subsection{Initial Calyx Segmentation}
\label{subsec:initial coral segmentation}

\begin{figure}
    \centering
    \includegraphics[width=\columnwidth]{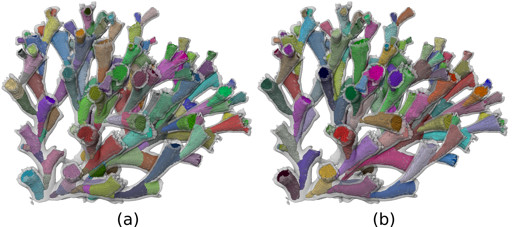}
    \caption{
        \emph{(a)}%
            ~The initial contour-tree-based instance segmentation of the polyp-cavity (calyx) segmentation has 153 instances and involves a trade-off between the over-segmentation of large, adult corallites and the under-segmentation of small, young corallites.
        \emph{(b)}%
            ~The final, manually corrected instance segmentation contains only 114 calices.
    }
    \label{fig:a2w contour tree segmentation}
\end{figure}

The first step in the processing of a 3D image containing coral colonies is the computation of the \emph{calyx (polyp cavity) segmentation}. 
This is accomplished with the ambient occlusion and ambient curvature-based method developed by Schmitt et al.~\cite{2021:schmitt}, which makes use of two observations.
Firstly, the cavities are more occluded than regions outside the skeleton~\cite{2016:Baum,2018:Titschack}.
Secondly, coral cavities are mostly convex on the outside and concave on the inside. 
Combining both information in a statistical model for background and foreground, i.e., cavity and non-cavity, gives a soft segmentation of the polyp cavities~(calices).
The final segmentation is then obtained by computing a minimal partition of the soft segmentation.
The result for a small colony fragment is shown in \hyperref[fig:teaser]{Figure~1b}.

In the next step, the initial calyx instance segmentation is computed by a classical contour-tree segmentation~\cite{1979:beucher} on the random-walk distance transform~\cite{2004:gorelick} of the calyx segmentation.
Although a classical Euclidean distance transform may be used as well, we share the findings of Baum et al.~\cite{2019:baum} that the random-walk distance transform yields slightly better results for the calyx instance segmentation.
The persistence value in the contour-tree segmentation is interactively chosen such that most of the small, young corals are segmented correctly.
This choice, however, comes with an over-segmentation of larger, adult corals. 
We find this tradeoff easier to work with than the other way around, since over-segmentations are easier to spot and correct.
The initial calyx instance segmentation of the same colony fragment as in \hyperref[fig:teaser]{Figure~1b} is shown in \hyperref[fig:a2w contour tree segmentation]{Figure~3a}.

At this point, the initial calyx instance segmentation can already be refined with the tools described in \autoref{subsec:refinement tools}.
The final result is depicted in \hyperref[fig:teaser]{Figure~3b}.

%% -- Automatic Colony-Tree Computation
\subsection{Automatic Colony-Tree Computation}
\label{subsec:automatic colony-tree computation}

\begin{figure*}
    \centering
    \includegraphics[width=\textwidth]{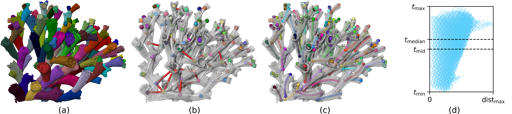}
    \caption{
        \emph{(a)} 
            The corallite mask is obtained by propagating the calyx instance segmentation onto the skeleton mask.
        \emph{(b)} 
            The region adjacency graph of the corallite instance segmentation (a) contains a vertex for each label (coral) and an edge for all labels sharing a border. 
            Highlighted are edges that are \emph{not} part of the colony tree and must be removed.
        \emph{(c)} 
            Parabolas are fitted to the calyx instance segmentation, allowing to orient each corallite from bottom to top.
        \emph{(d)} 
            The scatter plot shows the distribution of the voxel distances $ \lVert \gamma \left( t_i \right) - x_i \rVert $ to the fitted parabola of a calyx and its skew that is used to orient it.
    }
    \label{fig:preprocessing steps tree computation}
\end{figure*}

Among other information, we are particularly interested in the skeleton tree that describes the mother-daughter corallite relationship in a coral colony, that is, a graph with a vertex for each corallite and a directed edge going from the mother corallite to the daughter corallite.
Since corals reproduce asexually within a colony, the graph will be a tree.

The automatic tree computation starts by propagating the calyx instance segmentation, like the one shown in \hyperref[fig:a2w contour tree segmentation]{Figure~3b}, onto the corals skeleton.
The propagation algorithm assigns each voxel in the skeleton mask the label of the closest calyx.
The result is a corallite instance segmentation as the one shown in \hyperref[fig:preprocessing steps tree computation]{Figure~4a}.

Scanning through the corallite instance segmentation and tracking which instances touch each other yields a region adjacency graph (RAG) as shown in \hyperref[fig:preprocessing steps tree computation]{Figure~4b}.
This graph has one vertex for each corallite and one edge for each pair of neighbouring corallites. 
Unfortunately, the region adjacency graph is too large in the sense that it also contains edges between corallites that are adjacent but do not describe a proper mother-daughter relationship. 

In order to prune these superfluous edges, we first orient the corallites.
We start by fitting a parabola to the point cloud given by all voxels occupied by the corresponding calyx.
Let $ \mathcal{V} \subset \mathbb{R}^3 $ denote the point cloud of voxels belonging to a calyx.
Then we seek to find a rotation matrix $ R \in O(3) $, an anchor point $ a \in \mathbb{R}^3 $, the curvature of the parabola $ \alpha \in \mathbb{R}_{\geq 0} $, as well as the curve parameters $ t_x \in \mathbb{R} $ for all $ x \in \mathcal{V} $, such that the distance between the spatial parabola
$$
    \gamma(t) 
    = R \begin{pmatrix}
        t \\ \alpha t^2 \\ 0
    \end{pmatrix}
    + a 
$$
and the calyx point cloud is minimized. In more formal terms:
$$
    \min_{\substack{
        R \in O(3) \\
        a \in \mathbb{R}^3 \\
        \alpha \in \mathbb{R} \\
        t_x \in \mathbb{R}^3
    }}
    \sum_{x \in \mathcal{V}} \left\lVert \gamma(t_x) - x \right\rVert_2^2.
$$
In short, the parameters are initialized by first computing a principal component analysis (PCA) and projecting all voxels onto the first two principal axes.
The rotation matrix coincides, up to the sign of the last column, with the projection matrix of the PCA, while the anchor and curve parameters are found by linear regression.
The parameters are then refined by alternating the minimization with respect to the parabola parameters $ t_x $ and all other parameters.
\hyperref[fig:preprocessing steps tree computation]{Figure~4c} shows the parabola fit for the A2W specimen.

The parabola fit is finalized by swapping the start $ t_\text{min} = \min_{x \in \mathcal{V}} t_x $ and end $ t_\text{max} = \max_{x \in \mathcal{V}} t_x $, if necessary, such that more points in the point cloud project onto the second half of the parabola, i.e.,
$$
    \underbrace{\left\lvert \{ t_x : t_x < t_\text{mid} \right\} \rvert}_\text{number of points in first half}
    \leq 
    \underbrace{\left\lvert \{ t_x : t_x \geq t_\text{mid} \right\} \rvert}_\text{number of points in second half},
$$
with $ t_\text{mid} = \frac{t_\text{min} + t_\text{max}}{2} $. 
This condition is inspired by the observation that the coral grows in breadth.
\hyperref[fig:preprocessing steps tree computation]{Figure~4d} shows the distribution of the parameters $ t_x $ and the distance of the voxel $ x $ to the parabola for a single calyx.

Based on the oriented parabola, a simple measure describing whether a voxel is closer to the bottom or top of a calyx is defined as 
$$  
    h(x) 
    = \begin{cases}
        0   &, \quad\text{if } t_x < t_\text{min} \\
        \frac{t_x - t_\text{min}}{t_\text{max} - t_\text{min}}  &, \quad\text{if } t_\text{min} \leq t_x \leq t_\text{max} \\
        1   &, \quad\text{if } t_x > t_\text{max} \\
    \end{cases},
$$
where $ t_x $ again minimizes the distance of $ x $ to the parabola, i.e.,
$$
    t_x
    = \mathrm{argmin}_{t \in \mathbb{R}} \lVert \gamma(t) - x \rVert_2^2.
$$

An edge $ e = (A, B) $ connecting two calices \emph{A} and \emph{B} in the region adjacency graph is now oriented such that
\begin{equation}
    \label{eqn:edge orientation criterion}
    h_A \left( x_A \right) \geq h_B \left( x_B \right),
\end{equation}
where $ x_A \in V_A $ is the closest point in calyx A to calyx B, and $ x_B \in V_B $ is analogously the closest point in calyx B to calyx A:
$$
    \lVert x_A - x_B \rVert_2
    = \text{dist}(V_A, V_B).
$$
These points are easily obtained by tracking the touching points during the corallite instance segmentation.
The condition \autoref{eqn:edge orientation criterion} orients the edges such that they start at the top of the source corallite (mother) and end at the bottom of the target corallite (daughter).

Most of the remaining superfluous edges connect daughters of the same mother corallite, giving rise to the following configuration:
\begin{center}
    \begin {tikzpicture}[->,shorten >=2pt,auto,node distance=2cm,thick,main node/.style={circle,draw,font=\sffamily\Large\bfseries}]    
        \node[main node] (1) [minimum size=1cm]{$ \text{A} $};
        \node[main node] (2) [minimum size=1cm, below left of=1] {$ \text{B}_1 $};
        \node[main node] (3) [minimum size=1cm, below right of=1] {$ \text{B}_2 $};
        
        \draw (1) -- (2);
        \draw (1) -- (3);
        \draw[dashed,color=red] (2) -- (3);        
    \end{tikzpicture}
\end{center}
A simplification procedure prunes the edge between the daughters $ \text{B}_1 $ and $ \text{B}_2 $.

The pruned graph already gives a good estimate for the skeleton tree.
However, it may still contain edges for \emph{secondary joints}.
Formally, secondary joints are constituted by the subset of edges within the region adjacency graph that are not part of the skeleton tree.

\begin{figure}
    \centering
    \includegraphics[width=\columnwidth]{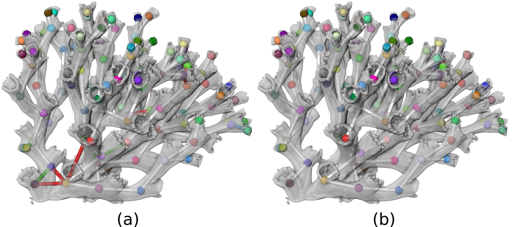}
    \caption{
        \emph{(a)} 
            After orienting the edges and pruning daughter-daughter relationships, the graph is much cleaner compared to \hyperref[fig:preprocessing steps tree computation]{Figure 4b} but still contains some wrong edges, highlighted in red, and wrongly oriented edges, highlighted in green.
            In total, it has 115 edges.
        \emph{(b)} 
            The final result after manual refinement by the user is a graph consisting of five trees, each describing the ancestry in a single colony, and has 109 edges.
    }
    \label{fig:a2w initial and final graph}
\end{figure}

The automatically computed skeleton graph and the final skeleton tree, after manual refinement of the skeleton graph, are shown in \autoref{fig:a2w initial and final graph}.

%% -- Refinement Tools
\subsection{Refinement Tools}
\label{subsec:refinement tools}

\begin{figure}
    \centering
    \includegraphics[width=\columnwidth]{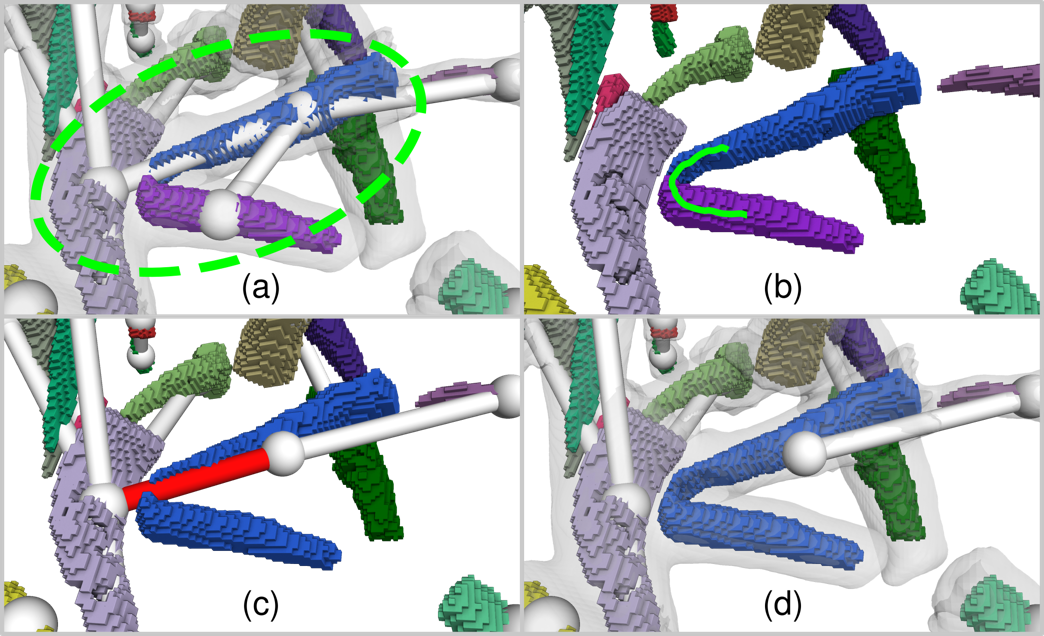}
    \caption{
        The images show the workflow of \textbf{correcting an over-segmentation} obtained by the initial contour-tree segmentation.
        \emph{(a)}%
            ~The initial state is shown.
            The exceptionally strongly bent corallite is separated in two segments that are joined by an edge and an additional secondary joint that connects them to the coral on the left.
        \emph{(b)}%
            ~The corals are merged by marking the two segments with a small scribble, shown in green. 
        \emph{(c)}%
            ~The merge tool preserves outgoing edges, so that the secondary joint needs to be removed in an additional step by marking it.
        \emph{(d)}%
            ~The final result shows the correct coral segmentation and connectivity.
    }
    \label{fig:merge workflow}
\end{figure}

The initial instance segmentation of the colony into calices and the associated skeleton graph are not perfect, that is, they may still have errors which require manual intervention. 
\CoDA\ provides suitable tools for correcting over-segmentations, under-segmentations, wrongly oriented edges, as well as adding missing and removing superfluous edges.
All operations acting on the instance segmentation keep the skeleton graph in sync by inserting or removing edges and vertices when needed.

Over-segmentations are corrected with the merge tool as shown in \autoref{fig:merge workflow}.
The user marks the instances that should be merged by drawing a line over them.
The voxels in the touched connected components are then collected and assigned to a new, common label. 
Their vertices in the spatial graph are merged as well and internal edges are removed while keeping the edges to non-affected vertices.

Under-segmentations are corrected by drawing a cutting line across the instance that is too large.
We fit a plane to the line and the viewing direction, and assign each half-space of the connected component a new label ID, resulting in a clean, scissor-like cut into two segments.
The edges of the original calyx are distributed between the two new instances such that they maximize the orientation condition stated in \autoref{eqn:edge orientation criterion}.
Additionally, a new edge is inserted between the two new components.
If no orientation of the view allows a clear cut, then the merge and cut tools can be used multiple times until a satisfactory result is achieved.

The vertices of the graph can only be changed by merging, cutting or reassigning the labels of the instance segmentation.
With the help of graph editing tools, the user can add missing edges by selecting the source and target vertices in the 3D view.
Similarly, superfluous edges are removed by selecting the edge itself and deleting it (\hyperref[fig:merge workflow]{Figure~6c}) as part of the merge workflow, or by selecting the source and target vertices and disconnecting them.

From time to time, the heuristic described in \autoref{subsec:automatic colony-tree computation} yields incorrect orientations, especially for small corals and for calices where top and bottom are close by.
In those cases, the direction of the adjacent edges can be corrected by selecting and flipping them.

\subsection{Proofreading}
\label{subsec:proofreading}

Proofreading in \CoDA\ consists of two steps. 
The first step concerns the calyx instance segmentation itself.
Here, a major part of wrongly segmented calices are identified by looking at the feature plots in \CoDAGraph, e.g., for volume, surface area, length or the mean distance to the parabolas.
Once all outliers are taken care of, a step-through process presents the user with all coral instances one after the other.
Initially, all instances are marked as \emph{unseen}.
The user then either marks them as \emph{good} or corrects them with the proper refinement tools described in \autoref{subsec:refinement tools}.
After refinement, all affected calyx instances are again marked as \emph{unseen} and placed into the proofreading queue.

Problems with occlusions, for example in inner, dense areas of a colony, are taken care of by showing only the closest, adjacent corals by cropping to a centered region of interest (RoI).
Notably, the order in which the user steps through the instances is not random but visits the next, closest, unchecked calyx.
Thus larger jumps within the 3D volume image are avoided.
The camera of the view is aligned with the principal axes of the calyx to obtain a standardized view.

\begin{figure}
    \centering
    \includegraphics[width=\columnwidth]{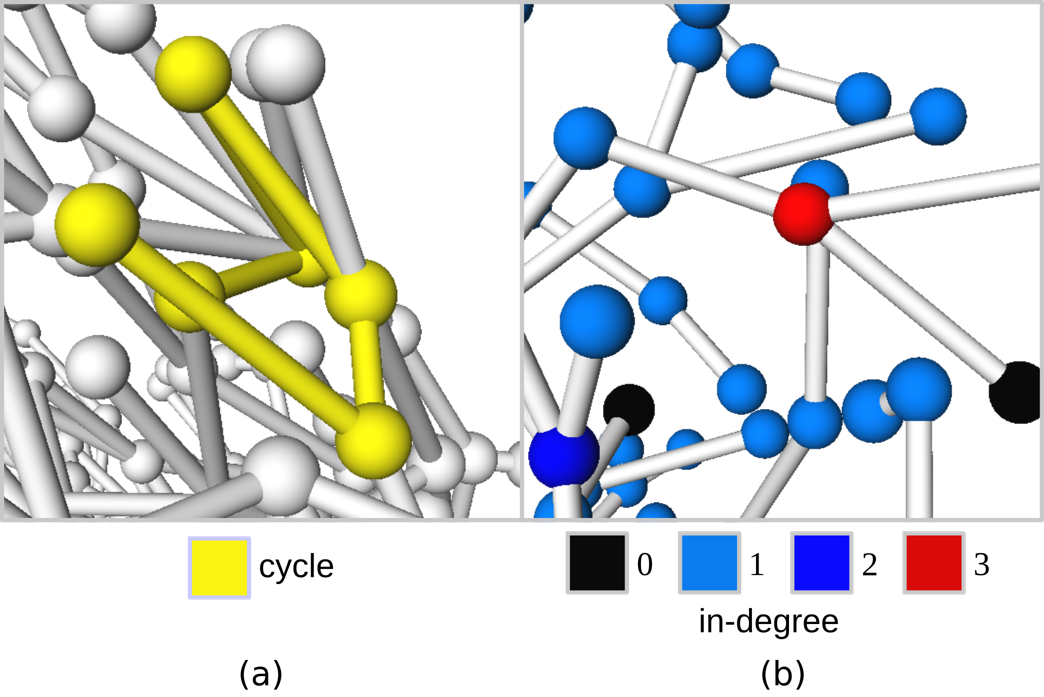}
    \caption{
        The initial graph computed with the method described in \autoref{subsec:automatic colony-tree computation} usually contains superfluous and wrongly oriented edges.
        Several visual hints are provided to easily identify these edges.
        \emph{(a)} 
            The shortest cycle in the graph is highlighted so that the user can quickly trace back superfluous edges, e.g., secondary joints, and remove them.
        \emph{(b)} 
            Also, each calyx/corallite can have at most one predecessor.
            This fact is used by the in-degree visual hint that shows the number of incoming edges a calyx/corallite (vertex) has.
    }
    \label{fig:visual hints spatial graph tree}
\end{figure}

The second proofreading step concerns the edges of the skeleton graph.
The cycle detection highlights the shortest cycle in the graph which can be traced to find and remove the edge that was caused by a secondary joint.
Each vertex must also have an in-degree of at most one since each calyx/corallite can have only one or no predecessor.
These two hints are the most helpful for identifying wrong edges and are shown in \autoref{fig:visual hints spatial graph tree}.
Other helpful features during this proofreading stage are the edge orientation criterion in \autoref{eqn:edge orientation criterion}, the budding angles, edge lengths or contact area widths.
It is important to note that the visual hints only work for calices and edges that are in the current segmentation, and that we cannot add visual hints for edges that are not there anymore, once they have been removed.
Thus, also for this step, we prefer that the initial instance segmentation is an over-segmentation and contains more calices as well as more edges due to secondary joints.

Similarly to the proofreading of the calyx instance segmentation, the user is provided with a step-by-step tool for proofreading all edges.
This guides the user through all vertices (calices).
The user marks a vertex as \emph{good} if all edges starting or ending in the vertex are correct and none are missing.

The proofreading state is just another attribute of the vertices and edges and can thus also be used as source for the color- and glyph-map.
This allows the user to easily see which vertices and edges are still marked as \emph{unseen} or as already been taken care of.

%% -- CoDAGraph
\section{Implementation}
\label{sec:implementation}

The whole proofreading, analysis and exploration framework is implemented within two different software stacks that communicate via a simple file-based mechanism.
The image processing and refinement tools are implemented in Amira~\cite{2005:Stalling} and are available as custom modules and extensions to the built-in 3D visual analysis tools.
Within Amira, \emph{aggregated} features are extracted for the individual calices and related corallites, as well as their relationships, and are exported to \CoDAGraph, a web application written in Python using the Bokeh~\cite{2018:bokeh} library.
\CoDAGraph\ extends Amira by a rich set of classical 2D visualizations like SPLOMs, scatter plots, histograms, and graph layouts.

%% -- Linking and Burshing between Amira and CoDAGraph
\subsection{Linking between Amira and CoDA.Graph}
\label{subsc:linking amira and codagraph}

\begin{figure*}
    \centering
    \includegraphics[width=\textwidth]{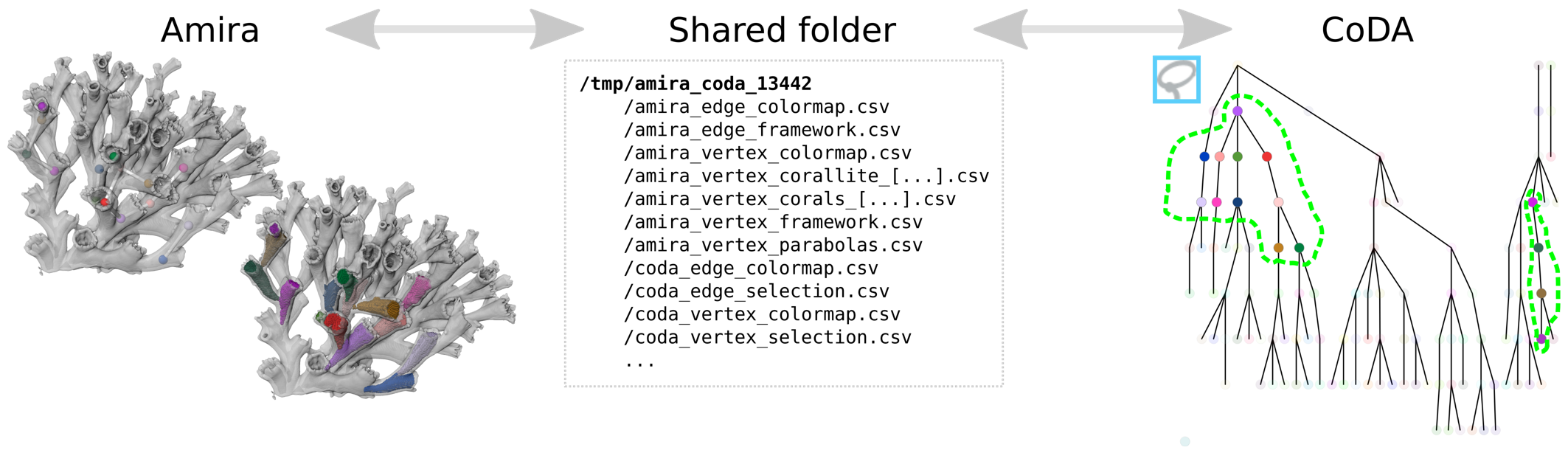}
    \caption{
        \CoDAGraph\ and Amira communicate over a simple file-based protocol with a canonical naming scheme.
        In this example, the \emph{lasso selection} in \CoDAGraph\ is propagated to Amira, resulting in a filtered view in which only the selected vertices are visible in the spatial graph visualization, as well as the selected calices in the volume visualization.
    }
    \label{fig:codagraph amira ipc}
\end{figure*}

\CoDAGraph\ and Amira are entirely decoupled and can be used independently. 
However, to facilitate an interactive and seamless exploration of the data, a simple communication protocol based on \texttt{.csv} files and a canonical naming scheme is implemented.

On startup, a temporary folder with a known prefix \texttt{amira\_coda\_\{random\_number\}/} is created by Amira. 
\CoDAGraph\ automatically detects folders with this prefix and watches their content.
Whenever a file in the shared folder is created, deleted or modified, the change is detected automatically by both Amira and \CoDA, which will trigger a reload of the respective spreadsheets.
To avoid race conditions or undefined behaviour, a spreadsheet is only ever written to by either Amira or \CoDA\ and read by the other, never by both. 
The origin of the data is again indicated with a prefix.
\begin{itemize}
    \item 
        \texttt{amira\_vertex\_\{name\}.csv} \\
        A spreadsheet with features exported from Amira in which each row contains data for one vertex (calyx/corallite).
        All vertex spreadsheets must have the same number of rows.
    \item 
        \texttt{amira\_edge\_\{name\}.csv} \\
        A spreadsheet with features exported from Amira in which each row contains data for one edge. 
        All edge spreadsheets must have the same number of rows.
        Additionally, at least one spreadsheet must have two columns with the index of the start and end vertices.
    \item 
        \texttt{amira\_vertex\_colormap.csv} \\
        A spreadsheet with the current color for every vertex (calyx/corallite). 
        The colormap can be used in \CoDAGraph\ so that the user is presented with the same color scheme in all views.
    \item 
        \texttt{amira\_edge\_colormap.csv} \\
        The same as \texttt{amira\_vertex\_colormap.csv} but here, each row contains the color for an edge.
    \item 
        \texttt{coda\_vertex\_selection.csv} \\
        A spreadsheet containing a single column in which each row indicates whether a vertex is currently selected in \CoDAGraph\ or not. 
        This information is used in Amira to show the currently selected calices/corallites in \CoDAGraph, enabling a seamless link-and-brush experience.
    \item
        \texttt{coda\_edge\_selection.csv} \\
        The same as \texttt{cora\_edge\_selection.csv} but each row indicates if an edge is currently selected in \CoDAGraph\ or not.
    \item 
        \texttt{coda\_vertex\_colormap.csv} \\
        The same as \texttt{amira\_vertex\_colormap.csv} but the colors are the ones displayed in \CoDAGraph.
    \item 
        \texttt{coda\_edge\_colormap.csv} \\
        The same as \texttt{amira\_edge\_colormap.csv} but the colors are the ones displayed in \CoDAGraph.
\end{itemize}
The data flow is visualized in \autoref{fig:codagraph amira ipc}.

At this point, it is worth mentioning that the naming scheme and choice of \texttt{.csv} files for the communication make it easy to use \CoDAGraph\ without Amira.
Similarly, an integration of \CoDAGraph\ into other applications is just a matter of adhering to the naming scheme within the shared folder.

%% -- Views and Panels
\subsection{Views and Panels}
\label{subsec:views and panels}

\CoDAGraph\ is implemented in Bokeh~\cite{2018:bokeh} and thus browser-based. 
It consists of up to three panels: A control panel and up to two viewer panels.

The leftmost panel is the control panel (\hyperref[fig:teaser]{Figure~1h}) and allows the user to change parameters such as the colormap, glyphmap, size and opacity for both edges and vertices. 
These visual settings are used inside all views, which gives a uniform impression.
Additional controls allow the user to choose the current visualization type for each view, e.g., a graph layout for the left and a scatter plot for the right view.
Each view type has its own settings which are also shown in the control panel.
For example, the scatter plot view has a control for choosing the column containing the features of the $ x $-axis and another one for choosing the features on the $ y $-axis.
\CoDAGraph\ aims to be as simple as possible and thus tries to reduce the number of controls available to the user or use sensible presets whenever possible. 

All classical plots are available, that is, \emph{scatter plots}, \emph{histograms}, \emph{SPLOMs} and \emph{petal (wedge) plots}.
Additionally, \CoDAGraph\ provides a \emph{map view} with the option for different tile providers and locations of the datasets. 
A map view showing the location of the five datasets considered in this paper is shown in \autoref{fig:inter colony exploration}.
The \emph{graph view} shows a layout computed with any of the algorithms provided by the Graphviz~\cite{2002:ellson} library.
If the loaded graph is a tree, the view defaults to the \emph{dot} view, so that a dendrogram of the generations is shown. 

\CoDAGraph\ also provides functional views for \emph{PCA} and \emph{UMAP}~\cite{2018:mcinnes}, two of the most popular techniques for dimensionality reduction.
Upon opening the view, the user chooses the features subject to the dimensionality reduction and views the result directly in a SPLOM.

The raw data can be viewed with the \emph{spreadsheet} view (\hyperref[fig:teaser]{Figure~1h}) and simple statistics like the data range, standard variance and quantiles are available in the \emph{statistics} view.

%% -- Exploration and Selection Tools
\subsection{Exploration and Selection Tools}
\label{subsec:exploration and selection tools}

\begin{figure*}
    \centering
    \includegraphics[width=\textwidth]{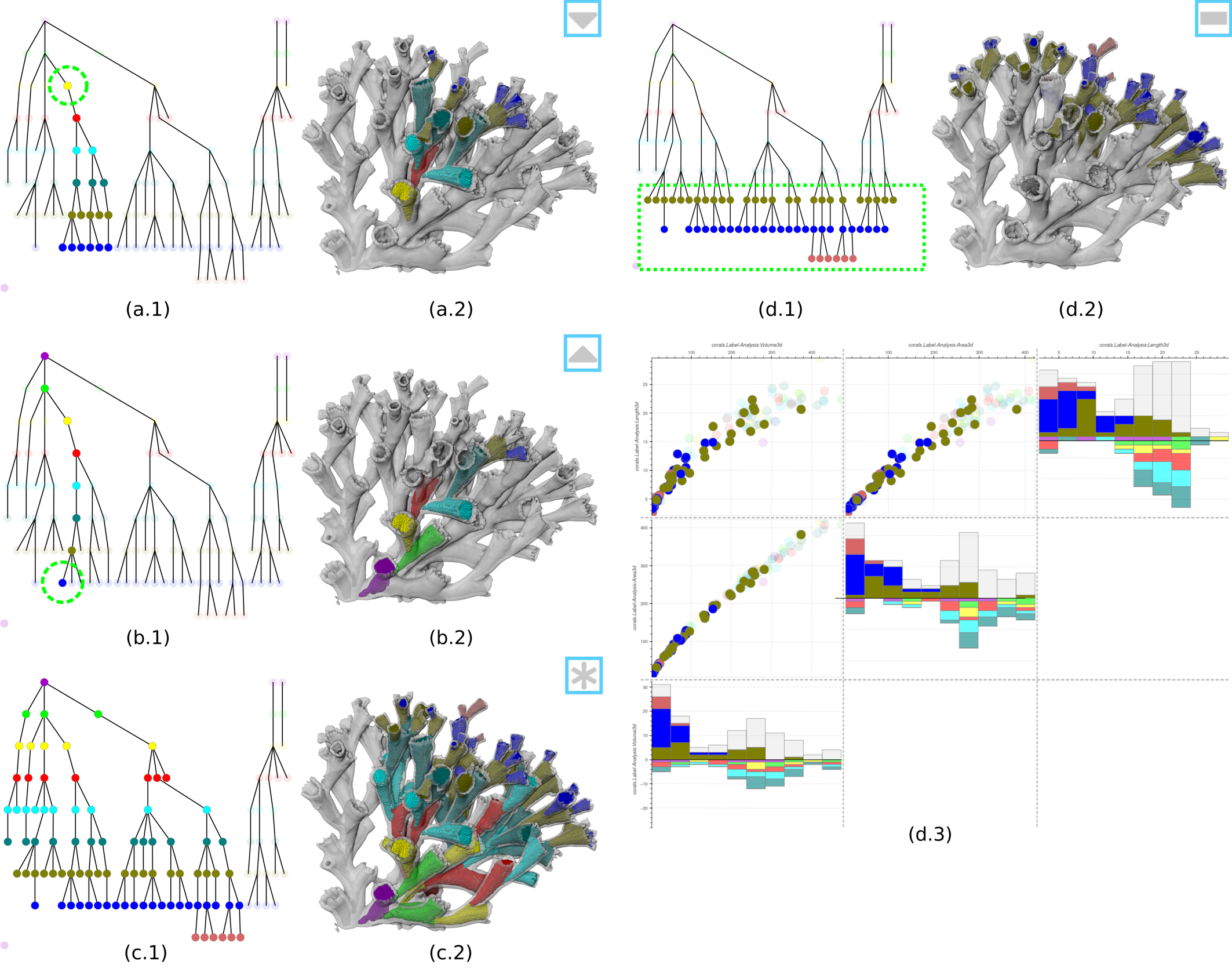}
    \caption{
        Depicted are several selections available in \CoDAGraph\ and their linked 3D views. 
        The calices are colored by their generation and the dendrogram is computed with the \emph{dot} graph layout. 
        \emph{(a.1, a.2)}%
            ~The \textbf{descendant selection tool} selects the whole sub-tree of a vertex, i.e., all descendants of the corallite represented by the vertex.
        \emph{(b.1, b.2)}
            The \textbf{ancestor selection tool} selects all ancestors of a single corallite. 
            Here, the user clicked onto the circled, blue vertex. 
            The path of the ancestors through the colony is clearly visible in the filtered 3D view.
        \emph{(c.1, c2)}%
            ~The largest of three colonies present in this specimen was selected using the \textbf{connected component} tool by clicking on one of the points belonging to the connected component.
        \emph{(d.1, d.2)}%
            ~The image shows a selection in \CoDAGraph\ with the \textbf{box selection tool}. 
            Due to the graph layout, the box selection tool is perfectly suited for selecting several generations at once. 
            Here, the selection box is coloured in green and encapsulates the youngest three generations of all colonies.
        \emph{(d.3)}%
            ~All views are linked, allowing the user to explore a hypothesis about the data in real-time. 
            The SPLOM views show the volume, area and length of the calices. 
            All corals that are not in the selection above \emph{(d.1)} are muted, eventually confirming the hypothesis that the youngest corals are not fully grown yet and will need up to four generations to mature.
    }
    \label{fig:cora selection tools}
\end{figure*}

Selections can be made in any of the \CoDAGraph\ views and will be propagated to Amira.
Bokeh comes with several useful selection tools out of the box, for example, the lasso, box and tap selection tools, which already allow for a freehand, unconstrained selection in all views.

For the ontogenetic analysis and exploration of the spatial growth patterns of the coral colonies, we implemented custom tools in the graph view.
The \emph{descendant} selection tool selects the sub-tree containing all descendants of the selected calyx/corallite;
the \emph{ancestor} selection tool selects all ancestors of a single corallite;
and the \emph{connected component} selection tool selects all corallites within a connected component, which are all calices/corallites within the same colony fragment.
These tools allow the exploration of a corallite's ancestry within a specimen and to focus on a single colony fragment within a dataset.
An overview of the selection tools is shown in \autoref{fig:cora selection tools}.

\section{Case studies}
\label{sec:case studies}

In order to showcase the suitability of \CoDA\ for the analysis of coral colonies, case studies are presented using the five specimens described in \autoref{subsec:data}.
The chosen samples differ in size, corallite number, and corallite and colony morphology, so that the study covers many use cases.

% -- Segmentation Case Study --
\subsection{Segmentation}
\label{subsec:case study segmentation}

\begin{table*}
    \centering
    \caption{
        For each of the specimens investigated in this paper, this table summarizes a lot of information about their processing and the specimens themselves.
        The table shows the species, scanning resolution and some simple, yet meaningful statistics extracted from the skeleton graph, e.g., the total number of colony fragments, calices/corallites and number of buddings.
        Listed are also the changes in the colony skeleton graph at different processing stages.
        The number of vertices coincides with the number of calices/corallites in each specimen.
        The pruning step from the RAG to the initial colony skeleton graph computed with the method described in \autoref{subsec:initial coral segmentation} greatly reduces the number of wrong edges.
        For correct edges, i.e., edges that are in the final instance segmentation, the automatic orientation produces mainly the correct orientation and fails only for very few edges.
    }
    \begin{tabular}{cccccc}
        \toprule             
            \textbf{Sample}
            & \textbf{\makecell{(I) A2W}} 
            & \textbf{\makecell{(II) C1W}} 
            & \textbf{\makecell{(III) SaM-ID43148}} 
            & \textbf{\makecell{(IV) GeoB127471-1}} 
            & \textbf{\makecell{(V) NIWA-148046}} 
        \\
            \textbf{Species}
            & \textit{Lophelia pertusa}
            & \textit{Lophelia pertusa}
            & \textit{Lophelia pertusa}
            & \textit{Madrepora oculata}
            & \textit{Goniocorella dumosa}
        \\
        \midrule
            \textbf{\makecell{Voxels}}
            & $ 257 \times 237 \times 392 $
            & $ 275 \times 145 \times 276 $
            & $ 1024 \times 729 \times 1215 $
            & $ 279 \times 115 \times 335 $
            & $ 829 \times 455 \times 317 $
        \\
            \makecell{\textbf{Resolution} $ [\text{mm}^3] $}
            & $ 0.351 \times 0.351 \times 0.3 $
            & $ 0.351 \times 0.351 \times 0.3 $
            & $ 0.3125 \times 0.3125 \times 0.3 $
            & $ 0.351 \times 0.351 \times 0.3 $
            % TODO: @Jürgen, did you use another scanner for these samples? The resolution is better.
            & $ 0.146 \times 0.146 \times 0.3 $
        \\
        \midrule
            \textbf{\makecell{Colony fragments}}
            & 5
            & 1
            & 67
            & 1
            & 9
        \\
            \textbf{\makecell{Calices/Corallites}}
            & 114
            & 30
            & 1725 
            & 66
            & 262
        \\
        \midrule
            \textbf{\makecell{Buddings}}
            &
            &
            &
            &
            &
        \\
            \makecell{Median}
            & 1
            & 1
            & 1
            & 1
            & 0
        \\
            \makecell{75\% Quantile}
            & 2
            & 2
            & 2
            & 2
            & 1
        \\
            \makecell{Maximum}
            & 4
            & 4
            & 7
            & 3
            & 20
        \\
        \midrule
            \textbf{\makecell{Edges}} 
            &
            &
            &
            &
            &
        \\
            \makecell{RAG}
            & 148
            & 36
            & 3592
            & 86
            & 409
        \\
            \makecell{Initialisation}
            & 115
            & 29
            & 2152
            & 67
            & 361
        \\
            \makecell{Final}
            & 109
            & 29
            & 1658
            & 65
            & 253
        \\
        \midrule
            \makecell{\textbf{RAG $ \rightarrow $ Initialisation} \\ \textit{(automatic)}}
            & 
            &
            &
            &
            &
        \\
            \makecell{Computation time}
            & 25s
            & 9s
            & 420s
            & 5s
            & 86s
        \\  
            \makecell{Unchanged edges}
            & 68
            & 0
            & 734
            & 29
            & 218
        \\  
            \makecell{Pruned edges}
            & 32
            & 7
            & 1440
            & 19
            & 48
        \\
        \midrule
            \makecell{\textbf{Initialisation $ \rightarrow $ Final} \\ \textit{(manual)}}
            & 
            & 
            & 
            & 
            & 
        \\  
            \makecell{Unchanged edges}
            & 107
            & 29
            & 1378
            & 55
            & 172
        \\
            \makecell{Removed edges}
            & 8
            & 0
            & 711
            & 9
            & 125
        \\
            \makecell{Added edges}
            & 2
            & 0
            & 217
            & 7
            & 17
        \\
            \makecell{Flipped edges}
            & 0
            & 0
            & 63
            & 3
            & 64
        \\
        \bottomrule
    \end{tabular}
    \label{tab:num edges num vertices benchmark}
\end{table*}

% - C1W & GeoB127471-1 -
\begin{figure*}
    \centering
    \includegraphics[width=\columnwidth]{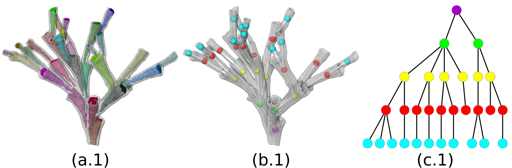}
    \includegraphics[width=\columnwidth]{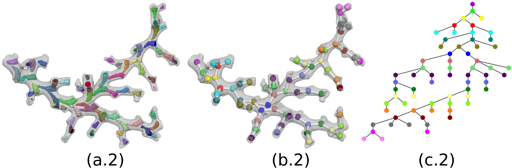}
    \caption{
        The segmentation results are shown for the \textit{Lophelia pertusa} \textit{C1W (II)} specimen on the left side and for the \textit{Madrepora oculata} \textit{GeoB127471-1 (IV)} specimen on the right side.
        Both consist of a single colony with \textit{C1W (II)} having 30 calices/corallites and \textit{GeoB127471-1 (IV)} having 66 calices/corallites.
        The calyx instance segmentation is shown in~\emph{(a.1, a.2)}; 
        the colony skeleton tree, colored by the generation, in~\emph{(b.1, b.2)}; 
        and the corresponding dendrogram in~\emph{(c.1, c.2)}.
    }
    \label{fig:c1w fig:GeoB127471-1}
\end{figure*}

% -- NIWA-148046
\begin{figure}
    \centering    
    \includegraphics[width=\columnwidth]{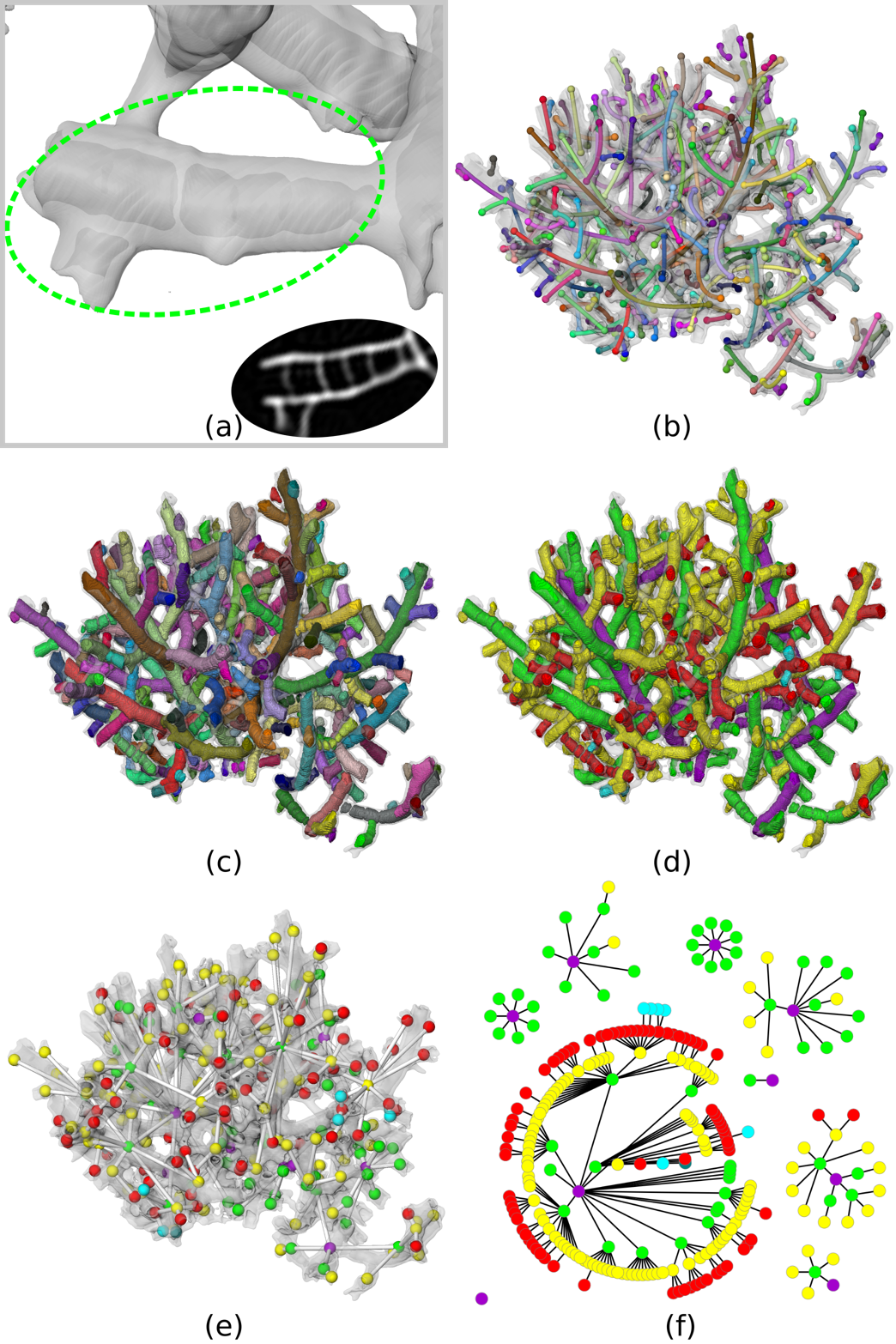}
    \caption{
        The processing of the \textit{Goniocorella dumosa} \emph{NIWA-148046 (V)} specimen is complicated by its completely different inner structure of the colony to other reef-forming stony corals such as \textit{Lophelia pertusa} and \textit{Madrepora oculata}.
        \emph{(a)}%
            ~The corallites of \emph{Goniocorella dumosa} are tube-shaped and build intermediate floors (dissepiments) in regular intervals, leading to an over-segmentation in the contour-tree segmentation.
        \emph{(b)}%
            ~The parabola fitting procedure also works for the large super-segments incorporating all segments belonging to a single corallite.
        \emph{(c)}%
            ~The image shows a rendering of all 262 calices that belong to the 9 colony fragments present in the specimen.
        \emph{(d, e)}%
            ~The images depict the calices colored by their generation and the underlying skeleton trees.
        \emph{(f)}%
            ~The plot shows an abstract graph layout of the skeleton trees computed with the \emph{two-pi} \cite{1997:wills-twopi} layout algorithm.
    }
    \label{fig:NIWA-148046}
\end{figure}

% -- SaM-ID43148
\begin{figure*}
    \centering    
    \includegraphics[width=\textwidth]{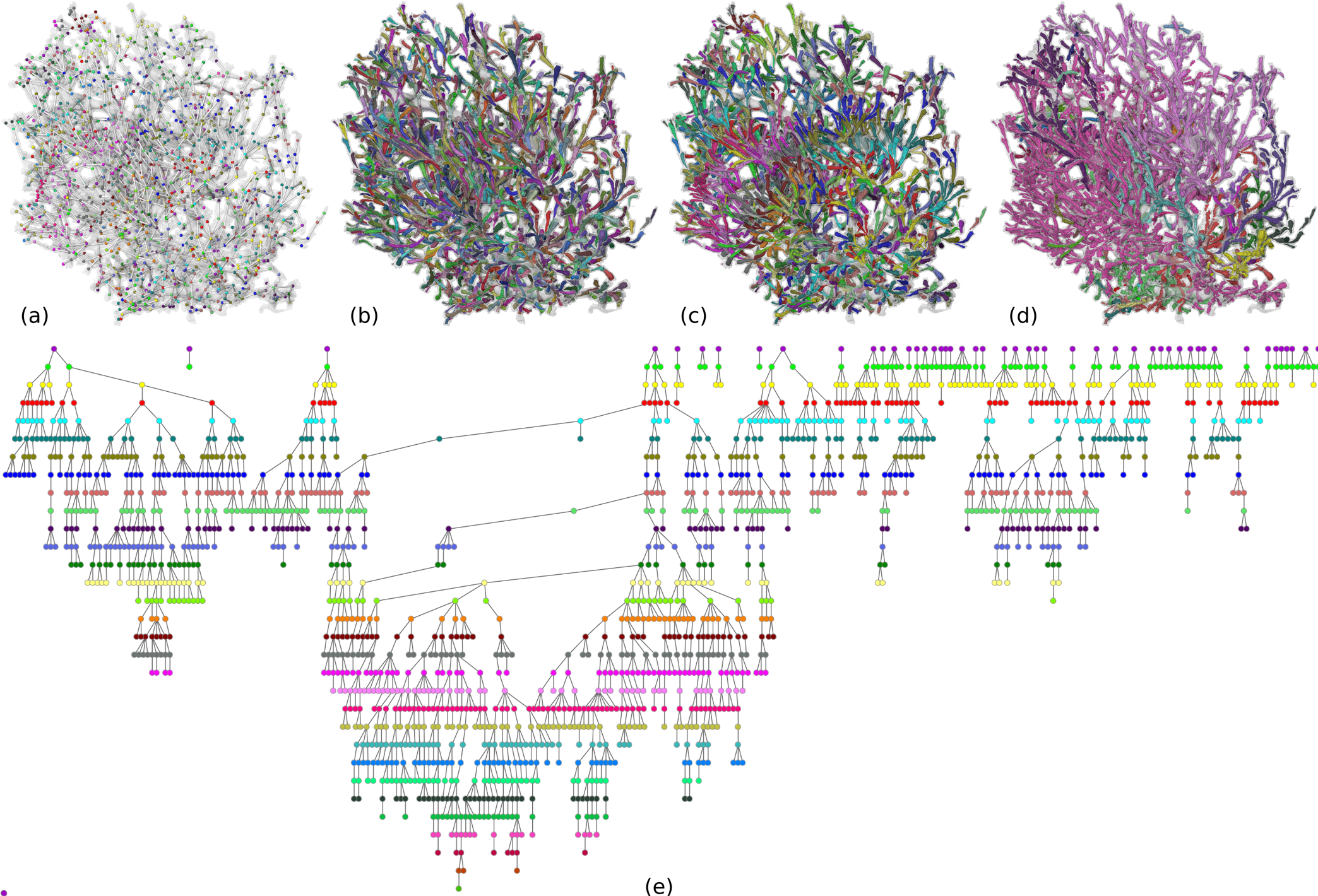}
    \caption{
        The images show the results for the \textit{Lophelia pertusa} \emph{SaM-ID43148 (III)} specimen, the largest one in our case studies made up of in total 66 colony fragments and 1724 calices/corallites.
        \emph{(a)}%
            ~shows the final skeleton graph;
        \emph{(b)}%
            ~shows the calyx instance segmentation;
        \emph{(c)}%
            ~shows the calices colored by their generation; 
            and
        \emph{(d)}%
            ~shows the corals colored by their connected component, i.e., the colony fragment they belong to;
        \emph{(e)}%
            ~shows the dendrograms computed with the \emph{dot} graph layout algorithm implemented in \CoDAGraph\ using the Graphviz \cite{2002:ellson} library.
    }
    \label{fig:SaM-ID43148 basic}
\end{figure*}

The initial instance segmentation was carried out similarly for all specimens and follows the workflow and recommendations from \autoref{sec:segmentation workflow}.
Separating a segment into two segments may require multiple cuts and subsequent merges since it is sometimes hard to find a cutting plane that separates both segments clearly, while merging two segments only requires a scribble roughly drawn on top of them.
Additionally, we find over-segmentations easier to spot than under-segmentations.
Two segments differ in color, while an under-segmented area is only recognizable by its shape or as an outlier in one of the morphological feature plots.

The comparatively small specimens of \textit{Lophelia pertusa}, \emph{A2W (I)}, \emph{C1W (II)}, and \textit{Madrepora oculata}, \emph{GeoB12747-1 (IV)}, were processed almost fully automatically; the whole process from calyx segmentation to the final colony-tree and instance segmentation took no longer than an hour each.
These specimens contain 114, 30 and 64 individual calices/corallites, respectively.
The fact that these specimens only contain a few or no secondary joints simplified the processing and manual interaction.
Due to their small size and morphology, the calices/corallites do not occlude others within the colony and it was possible to obtain a perfect segmentation without using the RoI centering and cropping tool in the proofreader.

The instance segmentation, skeleton-tree computation and proofreading of the largest sample, \emph{SaM-ID43148 (III)}, which motivated the development of \CoDA, took longer than every other specimen with roughly two working weeks.
After the initial, contour-tree-based calyx instance segmentation, we used \CoDAGraph\ to filter small instances with less than $ 0.5\mathrm{mm}^3 $ and removed them.
Such small cavities belong to small holes in the ambient occlusion mask that is used to obtain the polyp-cavity segmentation on which the calyx instance segmentation is based on, or artefacts due to the scanning resolution.
Even if they belong to young corallites, they are too small to be part of a meaningful morphological analysis, so that they would be discarded from further analysis anyway.
Next, we examined the initial calyx instance segmentation with the 3D visualizations and corrected all segments that we could spot at first glance.
Afterwards, we switched to the proofreading tool for the correction and validation of the calyx instance segmentation.
Especially the canonical view with the automatically centered RoI proved to be useful in the inner, dense areas of the specimen.
Notably, the calyx segmentation contained many tubes of \emph{Eunice} worms, which had to be removed.
Using the proofread instance segmentation, we computed the initial colony skeleton graph and pruned secondary joints with help of the cycle detection tool, of which \emph{SaM-ID43148 (III)} contains many.
A second proofreading was carried out, in which a few errors that we missed earlier were corrected, both in the calyx instance segmentation, as well as the colony-tree graph.
The final instance segmentation contains 1724 individual corals, 1658 edges and 66 colony fragments.

The processing of the \textit{Goniocorella dumosa} specimen \emph{NIWA-148046 (V)} proved to be more involved than the others due to its fundamental different morphology.
Each corallite of this species has intermediate floors (dissepiments) and budding occurs multiple times during corallite growth.
Hence, a single corallite corresponds to multiple calyx subcavities.
\hyperref[fig:NIWA-148046]{Figure~11a} shows an iso-surface rendering and a slice of the original CT intensity image of such a region.
In addition to the unique growth pattern, the calyx subcavities are more cylindrical in shape, other than \emph{Lophelia pertusa} or \textit{Madrepora oculata}, which are more conic.
Thus, they do not exhibit a strong anisotropy in growth direction which hampered the automatic computation of their orientation and the subsequent automatic estimation of the coral colony tree.
Contrary to the other corals, here, we used the Euclidean distance transform as scalar field for the contour-tree segmentation, which gave slightly better results due to the more cylindrical shape and already good separated calyx subcavities due to the intermediate floors.
The major bottleneck in the proofreading and refinement of this specimen was the merging of the many calyx subcavities belonging to the same corallites.
From start to end, the first refinement and proofreading process took roughly two work days and resulted in an instance segmentation with 262 corallites.

% -- Visual Exploration Case Study --
\subsection{Visual Exploration}
\label{subsec:case study visual exploration}

After finishing the segmentation of all specimens, we started the visual exploration by trying to confirm some basic hypotheses and questions.

The SPLOM plots of the volume and surface area of both calices and corallites colored by their generations showed us that the corals need a growth-period of up to four generations to reach their adult stage, visible in \hyperref[fig:cora selection tools]{Figure~9d.3} since they are predominantly smaller in size and length than the older generations.
After this period, morphological features are indistinguishable even in different branches of the same colony and between colony fragments in the same specimen.

The dendrograms also showed a skeleton tree that grows more in breadth rather than in depth for the \emph{Lophelia pertusa} specimens compared to the \emph{Madrepora oculata} specimen.
Taking a look at the statistics panel, we could also see that the maximum number of buds (three) in the \emph{Madrepora oculata} specimen is lower than the maximum number of buds (seven) in the \emph{Lophelia pertusa} specimens. 
Note that these are first impression on morphological differences between two species that must be confirmed by the analysis of many more specimens. 

In all cases, the colony-tree graph revealed more components, i.e., colony fragments, within a single specimen than identified by looking at the real specimens or the CT images. 

By coloring the calices depending on their colony fragment in the \emph{SaM-ID43148 (III)} sample, we can easily make out the principal directions of growth visually, as shown in \hyperref[fig:SaM-ID43148 basic]{Figure~12d}.
While younger generations tend to start growing in a flat area near the ocean floor into many directions, all later generations tend to follow the same direction towards the top left corner in the image.

\begin{figure*}
    \centering
    \includegraphics[width=\textwidth]{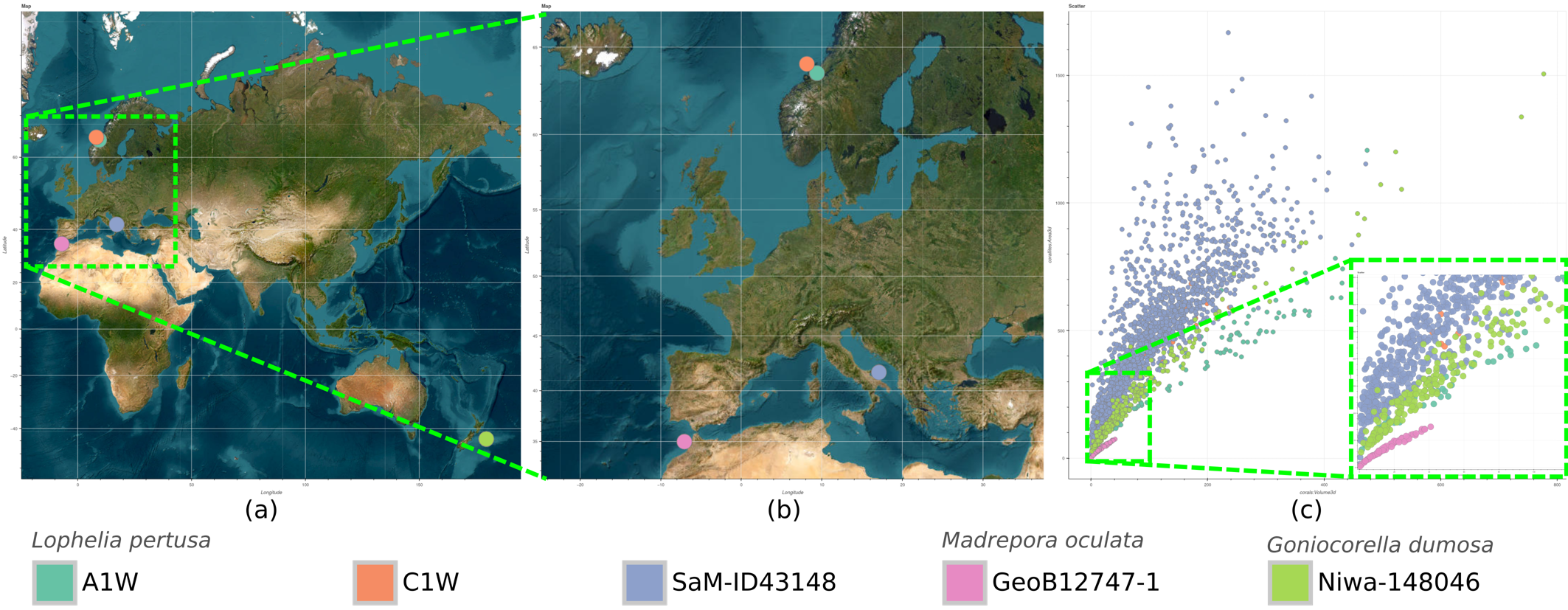}
    \caption{
        \CoDAGraph\ has built-in support for the analysis of multiple datasets at once by concatenating spreadsheets with the same name.
        \emph{(a, b)}%
            ~The images show the location of five different datasets on a world map and a zoomed view. 
            The location, i.e., latitude and longitude, are simply given as columns in one of the vertex-feature spreadsheets.
            % Copyright notice according to: https://developers.arcgis.com/documentation/mapping-apis-and-services/deployment/basemap-attribution/
            (Tiles~CC~Esri -- Source: Esri, i-cubed, USDA, USGS, AEX, GeoEye, Getmapping, Aerogrid, IGN, IGP, UPR-EGP, and the GIS User Community)
        \emph{(c)}%
            ~The scatter plot shows surface area and volume of all corals. 
            The color in \CoDAGraph\ is chosen based on the specimen ID so that the different locations are visually distinguishable.
    }
    \label{fig:inter colony exploration}
\end{figure*}

\CoDAGraph\ also allows the exploration of multiple colonies at once.
We use this feature of \CoDAGraph\ to compare the morphology and the ontogenetic development between different specimens/species.
All datasets are annotated with location attributes and their ID, which we use to compare the morphologies with respect to their species and location.
As shown in \autoref{fig:inter colony exploration}, the scatter plot of the calyx volume and calyx surface area reveal different growth patterns and shapes for different species.
The same plot also reveals that the correlation between volume and surface area depends on the species.

%% -- Performance --
\subsection{Performance}
\label{subsec:performance case study}

All analyses and explorations presented here were performed on a desktop PC with the following configuration: CPU: \textit{Intel Core i9-10920X} CPU; GPU \textit{Nvidia GeForce RTX 3090}, 24 GiB VRAM; Memory: 128 GiB DDR4.

The linking and brushing within \CoDAGraph\ works for all specimens without noticeable lags.
This is mainly due to the efficient implementation provided by \emph{Bokeh}, and the limited number of calices/corallites and thus rows in the spreadsheets.

The file IO required to pass data between Amira and \CoDAGraph\ is also not a bottleneck for the colony sizes considered here, since it happens inside temporary directories which are usually mounted directly in-memory.
Furthermore, the \texttt{.csv} file format is so simple, that loading and saving spreadsheets is not a concern.

The update of the 3D visualization in Amira however required a few seconds for the small corals like \emph{C1W (II)} and up to 20s for \emph{SaM-ID43148 (III)}, depending on the number of 3D visualizations that are shown at the same time, e.g., the filtered calyx instance segmentation overlayed with the skeleton-tree and an iso-surface rendering of the original CT scan. 
When the plugins in Amira detect a change inside the shared directory, the \emph{.csv} are loaded and parsed, followed by the generation of a new, filtered volume scalar field.
This scalar field is then uploaded to the graphics card before it is finally rendered.
After that, the volume can be explored at interactive frame rates.
Although the filtering could happen directly inside the shaders, making it only necessary to upload the indices of the selection to the graphics card, this would have required more effort and changes to Amira than just adding a new plugin.

%% -- Discussion --
\section{Discussion}
\label{sec:discussion}

Three-dimensional dendroidal structures are quite common in nature (e.g., trees, coral frameworks) and structure habitats, such as forests or coral reefs, increase their complexity which results in an increase of ecological niches and consequently enhanced biodiversity (e.g., \cite{2020:pygas, 2022:gamez}).
While the provided ecological niches and biodiversity in these habitats are quite well studied, the influence of the actual shape of the three-dimensional dendroidal organisms on the habitat complexity, the ecological niches and biodiversity are nearly unstudied.
One reason for this lack of knowledge is the predominantly qualitative or semi-quantitative description of these complex structures.
The herein presented \CoDA\ framework provides tools to assess such structures.

Currently, the presented segmentation pipeline and tools are optimised for dendroidal cold-water coral colonies with different corallite and colony morphologies. 
Even though the development process used only specimens of \textit{Lophelia pertusa}, \CoDA\ works well for two other cold-water coral species, i.e., \textit{Madrepora oculata} and \textit{Goniocorella dumosa}. 
However, the different corallite morphology of \textit{Goniocorella dumosa} with its tube-shaped corallites with multiple intermediate floors (dissepiments) ~\cite{1960:Squires} compared to the cone-shaped corallites of \textit{Lophelia pertusa} and \textit{Madrepora oculata}) clearly shows the sensitivity of the colony skeleton-tree construction on the calyx/corallite instance segmentation and the subsequent ontogenetic interpretation of the results. 
In \textit{Lophelia pertusa} and \textit{Madrepora oculata}, corallite budding occurs at a specific time of corallite development and usually is a single or double budding (maximal budding up to 7, see \autoref{tab:num edges num vertices benchmark}). 
Budding numbers greater than two occur predominantly in \textit{Goniocorella dumosa}. 
Its tube-shaped corallites exhibit up to $>20$ consecutive budding events (\autoref{fig:NIWA-148046}, \autoref{fig:NIWA-148046 dot layout}), which has fundamental importance for the ontogenetic interpretation of its colony skeleton tree. 
For an ontogenetic analysis of a colony skeleton tree of \textit{Goniocorella dumosa}, the calyx/corallite instance segmentation should not be based on the complete corallites (as shown herein) but on the calyx subvolumes enclosed by the intermediate floors (dissepiments; \hyperref[fig:NIWA-148046]{Figure~11a}).

After a meaningful calyx/corallite instance segmentation, the biggest challenge and prerequisite for the ontogenetic analysis of coral colonies is the correct reconstruction of the colony skeleton tree. 
A single wrong edge will cause the generation number in a whole branch to be wrong and multiple errors will accumulate fast.
With \CoDA, we now have a systematic approach and intuitive tools at hand for correcting and proofreading the colony skeleton tree so that the number of errors can be minimized.

Having proofread data for the calyx/corallite instance segmentations and methods to compute it in a timely manner enable us to gather enough training data to leverage machine learning methods, for example, a simple convolutional neural network like a U-Net~\cite{2015:ronneberger}, for learning the corallite borders.
The application of this may reduce the number of initial errors such that manual correction of errors will require less time.
Up until now, no sufficient training data had been available.
The coral colony datasets analyzed within this paper comprise almost 2200 corallites and their adjacencies, making it already a feasible training dataset for modern machine learning methods.

The initial automatic colony skeleton graph computation greatly reduces the amount of superfluous edges in the region adjacency graph.
However, some edges are falsely removed, mostly in dense areas within a colony with many secondary joints.
Often, these regions are interesting by themselves so that a closer look and manual introspection is required nonetheless.
Similarly to the proofread calyx/corallite instance segmentations, the now available colony skeleton trees can be used to bootstrap other methods for classifying the adjacency between corallites, e.g., deep learning-based methods, which will work directly on the original image data and will potentially be able to identify the corallite orientation and mother-daughter relationship as well as the secondary joints even more accurately.
Another option might be to employ graph neural networks for link prediction~\cite{zhang2018link}.

The parabolas fitted to the calices are not only useful for the colony skeleton tree computation, but provide new morphological features.
For example, the mean, standard deviation and skew of the distance distribution give a simple description for the anisotropy of a calyx, that is, the skew of the distribution will be larger for a cone-shaped calyx than for a tubular-shaped one.
Each parabola itself provides a robust estimate for the calyx (and corallite) centerline, its curvature provides a measure for the bending of the calyx/corallite, and the arc length provides a better estimate for the calyx' length than the one based on principal axes (PCA).

Although \CoDAGraph\ provides only minor contributions on its own, it combines many proven visualization techniques with a simple user interface and allows the exploration of cold-water corals in a novel, intuitive way. 
Being able to take a look at the region in the original volume image on which a data point in a plot is based on, may it be a classical measurement or otherwise computed feature, proved to be a very handy and convenient tool during the exploration and analysis of the coral colonies.
Especially the linked 3D visualization in Amira and the 2D graph-layout in \CoDAGraph\ were remarkably useful during the analysis of the corallite ancestry and provided a way to explore the relationships that would not have been possible within a single 2D or 3D visualization.

%% -- Conclusion and Future Work
\section{Conclusion and Future Work}
\label{sec:Conclusion and Future Work}

This study provides the technological basis to study complex three-dimensional dendroidal structures systematically. 
Exemplified on dendroid stony cold-water corals, \CoDA\ and \CoDAGraph\ are capable to disassemble fragments of dendroidal coral colonies into their principle building blocks (corallites) and construct an ontogenetic meaningful colony skeleton graph -- the basis for a quantitative and reproducible analysis of the corallite and coral-colony morphology. 
Based on this decomposition of coral colonies, an automated classification of colonial cold-water corals, the investigation of intra- and interspecies coral morphoplasticity, local or regional spatial preferences of the distribution of specific morphologies, environmental influences on the coral morphology and its influence of reef-forming capacity of specific morphologies become possible, previously hindered by the qualitative to semi-quantitative acquisition of morphological data of cold-water corals.
Our next goal is the statistical and morphological analysis of the most important reef-forming cold-water coral colonies to provide an automated CT-based species identification and investigate similarities and differences in their ontogenetic colony formation, corallite and coral colony morphology and their species-specific morphoplasticity. 

The preliminary results presented herein regarding the quantifiable morphological differences between species (\autoref{fig:inter colony exploration}), are already very promising. 
However, these analyses must currently be restricted to younger parts of colonies as older parts are impacted by bioerosion, especially by bioeroding sponges that produce large cavities that change the shape of the coral skeleton considerably (e.g., \cite{2005:beuck, 2007:beuck}). 
To overcome this restriction, we plan to use the data gathered from the analysis of intact coral colonies, like the ones presented in this paper, to develop and validate methods that deal with bioerosion-altered coral skeletons.
Furthermore, the software provides the first step to investigate cold-water coral deposits in more detail. 
So far, CT scans of sediment cores retrieved from cold-water coral reefs/mounds were only used to reconstruct the coral content, the coral clast size and the clast orientation (\cite{2014:Douarin, 2015:Titschack, 2016:titschack}). 
The incorporation of morphological analyses into these analyses following the approach of \CoDA\ will potentially allow the differentiation of coral species and identification of specific coral morphotypes, which will considerably improve the reconstruction of the temporal development of these cold-water coral reefs/mounds.

We make \CoDAGraph\ available to the public in the hope that it is not only useful for the visual analysis of cold-water coral colonies but also for other dendroid organisms and objects with an inherent tree-like structure.

%% -- Data and Code Availability --
\section{Data and Code Availability}
\label{sec:data and code availability}

The most recent version of \CoDAGraph\ is available at \url{https://github.com/zibamira/CTCoral_CoDA}.
The repository contains the documentation for the link and brush interface described in \autoref{subsc:linking amira and codagraph}, as well as some of the extracted features for the \emph{A2W (I)} specimen.

Additionally, we provide the source code for the Amira modules interfacing with \CoDAGraph\ at \url{https://github.com/zibamira/CTCoral_hxcoda}, the shared library for Amira 2024.core is also provided.

The original CT scans used in this paper as well as the final segmentation results and the spreadsheets with the computed features are available in the the PANGEA data repository.
\begin{enumerate}[(I)]
    \item 
        \emph{A2W} \\
        % \url{https://doi.pangaea.de/00.0000/PANGAEA.000000}
        \textit{Submitted}.
    \item 
        \emph{C1W} \\
        \url{https://doi.pangaea.de/10.1594/PANGAEA.947276}
    \item 
        \emph{SaM-ID43148} \\
        \url{https://doi.pangaea.de/10.1594/PANGAEA.947300}
    \item
        \emph{GeoB127471-1} \\
        \url{https://doi.pangaea.de/10.1594/PANGAEA.947334}
    \item 
        \emph{NIWA-148046} \\
        % \url{https://doi.pangaea.de/00.0000/PANGAEA.000000}
        \textit{Submitted}.
\end{enumerate}

%% -- Acknowledgements 
\section*{Acknowledgements}

Arne-Jörn Lemke and Christian Timann, Klinikum Bremen-Mitte, Gesundheit Nord (Bremen), are acknowledged for performing the CT scans and their support during the measurements. 
We thank Claudia Wienberg (MARUM) for providing access to the \emph{GeoB12738-1} specimen, and André Freiwald and Giovanni Sanna (both Senckenberg am Meer) for providing access to \emph{SaM-ID43148}.
Armin Form (GEOMAR) is acknowledged for providing CT scans of the specimens \emph{C1W} and \emph{A2W} (BMBF project BIOACID II (FKZ 03F0655A)).
Furthermore, we thank the nautical and scientific crews and especially the chief scientists of RV POSEIDON cruise POS455, PV PELAGIA cruise 64PE284, and RV METEOR cruise M70-1 during which the samples were retrieved.
Di Tracey and Sadie Mills (both from the National Institute of Water and Atmospheric Research (NIWA, New Zealand) and the NIWA Invertebrate Collection, are gratefully acknowledged for providing the \textit{Goniocorella dumosa} colony NIWA-148046. 
The specimen was collected by NIWA as part of their research project “Resilience Of deep-sea Benthic fauna to the Effects of Sedimentation” (ROBES) funded by the New Zealand Ministry of Business, Innovation and Employment (MBIE).
Furthermore, we thank Leiss Abdal Al for helping with the segmentation of the \emph{SaM-ID-43148} specimen.
The work in this paper is part of the project ``CTcoral – CyberTaxonomic Classification and Morphological Characterisation of Cold-Water Corals'' funded by the Deutsche Forschungsgemeinschaft (DFG, German Research Foundation; Ba5042/6-1; Ti706/6-1) as project 490665120.

%% -- Bibliography
\bibliographystyle{plain}
\bibliography{main.bib}

%% -- Author biographies
%
%   TODO: There is also \begin{IEEEbiography} if some of you want to include a picture of you.
%

% TODO: Sounds a little bit dry ...  :/ (And weird writing in third person about yourself.)
\begin{IEEEbiographynophoto}{Kira Schmitt}
\orcidlink{0009-0000-0385-0404} received her B.Sc.\ and M.Sc.\ degree in mathematics with a minor in computer science from the Technical University of Kaiserslautern (RPTU).
She worked at the Fraunhofer~Institute~for~Industrial~Mathematics during her studies and gained a strong interest in image processing and visualization.
Now, she develops methods for the automatic analysis of cold-water corals based on CT images as a full time researcher at the Zuse~Institute~Berlin as part of her Ph.D.\ thesis.
A project that combines her interests in image processing, nature and the ocean life. 
E-Mail: schmitt@zib.de. 
\end{IEEEbiographynophoto}
\begin{IEEEbiographynophoto}{Jürgen Titschack}
\orcidlink{0000-0001-9373-9688}  is carbonate sedimentologist at Marum - Center for Marine Environmental Science, University of Bremen specialised in non-tropical carbonate systems. He is interested in the integration of sedimentological, geochemical and palaeontological datasets of cold-water coral systems and island shelve carbonates to investigate their fate during climatic changes, and the development and application of computed tomography-based methodologies for the analysis of biogenic skeletons, shells, traces and, as well as within, sediment cores.
E-Mail:  jtitschack@marum.de.
\end{IEEEbiographynophoto}
\begin{IEEEbiographynophoto}{Daniel Baum}
\orcidlink{0000-0003-1550-7245}
studied computer science at Humboldt University Berlin (HUB), Germany, and the University of Edinburgh, Scotland.
He received his MS degree from HUB and his PhD degree from Freie Universit\"at Berlin, Germany.
During his PhD, he worked in the field of molecular visualization and similarity.
Later, his research interests shifted towards image analysis.
He is head of the research group Visual Data Analysis at Zuse Institute Berlin (ZIB), where he leads quite diverse projects from neurobiology to meteorology to the virtual unfolding of ancient written documents and the analysis of cold-water corals.
E-mail: baum@zib.de.
\end{IEEEbiographynophoto}

% Prepare the supplementary materials section
\vfill

\newpage
\renewcommand{\thepage}{S\arabic{page}}
\renewcommand{\thesection}{S\arabic{section}}
\renewcommand{\thetable}{S\arabic{table}}
\renewcommand{\thefigure}{S\arabic{figure}}
\setcounter{figure}{0}
\setcounter{page}{1}

\section*{Supplementary Materials}
\label{sec:supplementary}

\begin{figure*}
    \centering
    \includegraphics[width=\textwidth]{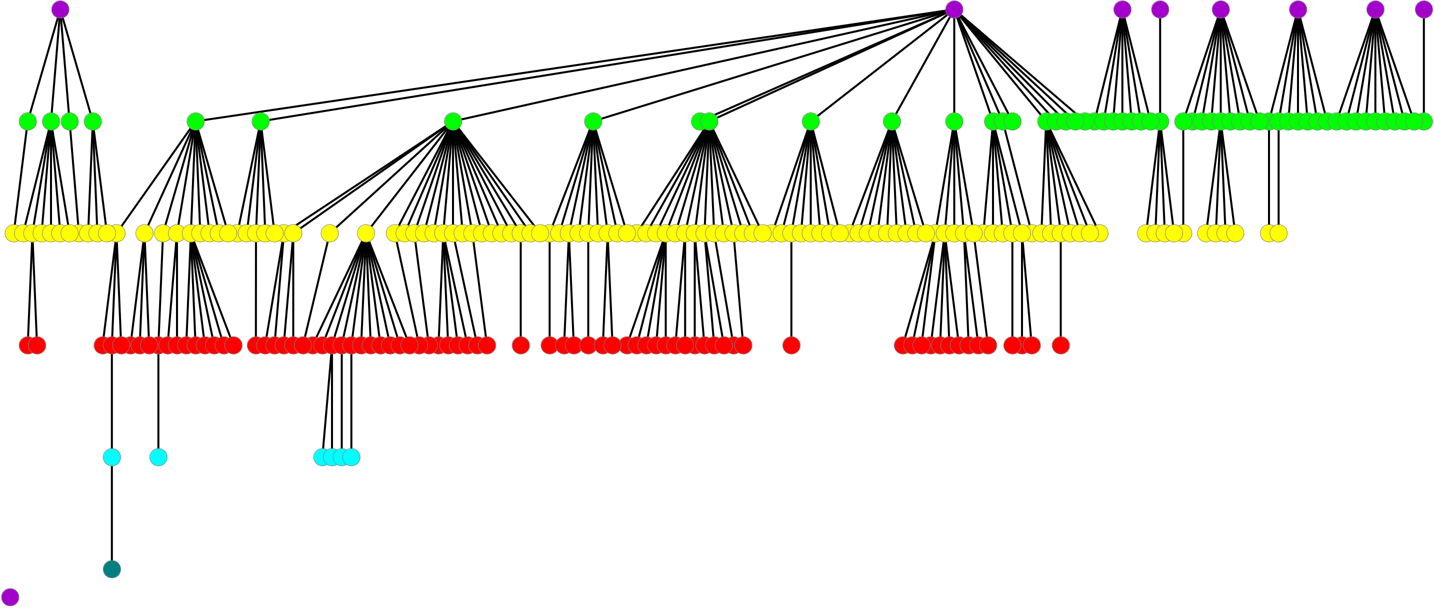}
    \caption{
        The image shows the dendrogram for the \emph{NIWA-148046} specimen, also shown in \autoref{fig:NIWA-148046}.
        For this species, the skeleton-tree does not correspond to the ontogenetic development, since it does not take into account the consecutive budding during corallite growth. 
        An instance segmentation based on the calyx subvolumes (cavities between intermediate floors (dissepiments)) might likely allow the construction of an ontogenetically meaningful skeleton tree.
    }
    \label{fig:NIWA-148046 dot layout}
\end{figure*}

\begin{figure}
    \centering
    \includegraphics[width=\columnwidth]{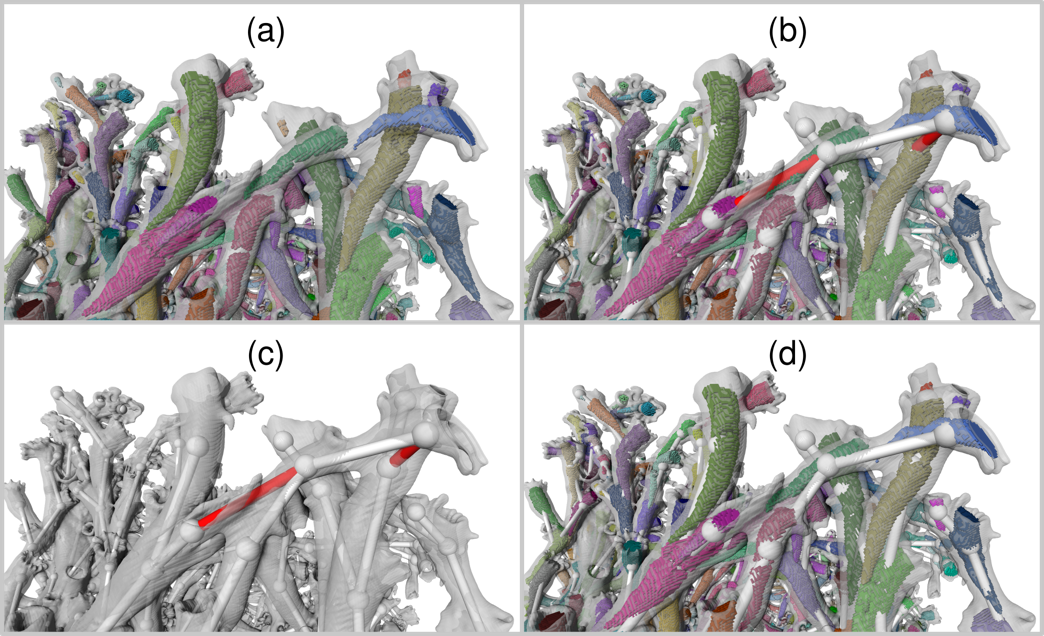}
    \caption{
        The images depict the workflow of \textbf{removing secondary joints}.
        \emph{(a)}%
            ~A region with multiple secondary joints, highlighted in red in (b) and (c). 
            Here, multiple coral branches are grown together.
        \emph{(b, c)}%
            ~The skeleton graph contains the secondary joints as superfluous edges (red) that must be removed. 
            The iso-surface rendering of the coral CT-scan is important for deciding whether an edge is a secondary joint or not. 
            The user may make use of different visualizations in this step. 
            Then, the secondary joints are marked and removed.
       \emph{(d)}%
            ~The image shows the final result without secondary joints.
    }
    \label{fig:secondary joints removal workflow}
\end{figure}

\begin{figure}
    \centering
    \includegraphics[width=\columnwidth]{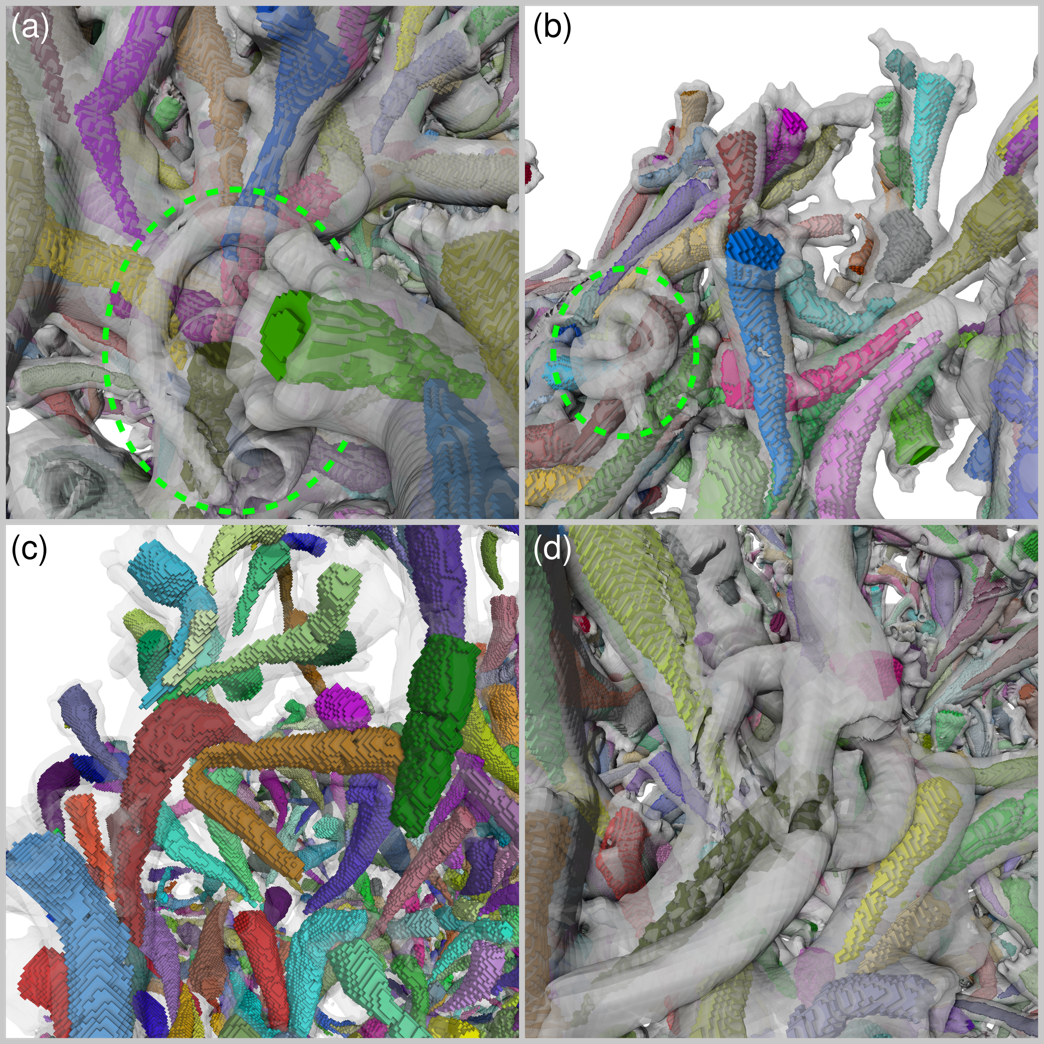}
    \caption{
        The images show some representative regions of the \emph{SaM-ID43148} specimen responsible for the difficulties involved in its analysis.
        The specimen contains many tubes belonging to \emph{Eunice} worms which have to be removed during the segmentation, one is highlighted in green in images \emph{(a)} and \emph{(b)}, and one is prominently shown in image \emph{(d)}.
        Some are very similar in shape to the \emph{Lophelia pertusa} corallites, making it hard to identify them automatic.
        The density of the corallites and the number of secondary joints impose additional difficulties onto the segmentation and proofreading process.
    }
    \label{fig:sam-id43148 impressions}
\end{figure}

\begin{figure*}
    \centering
    \includegraphics[width=\textwidth]{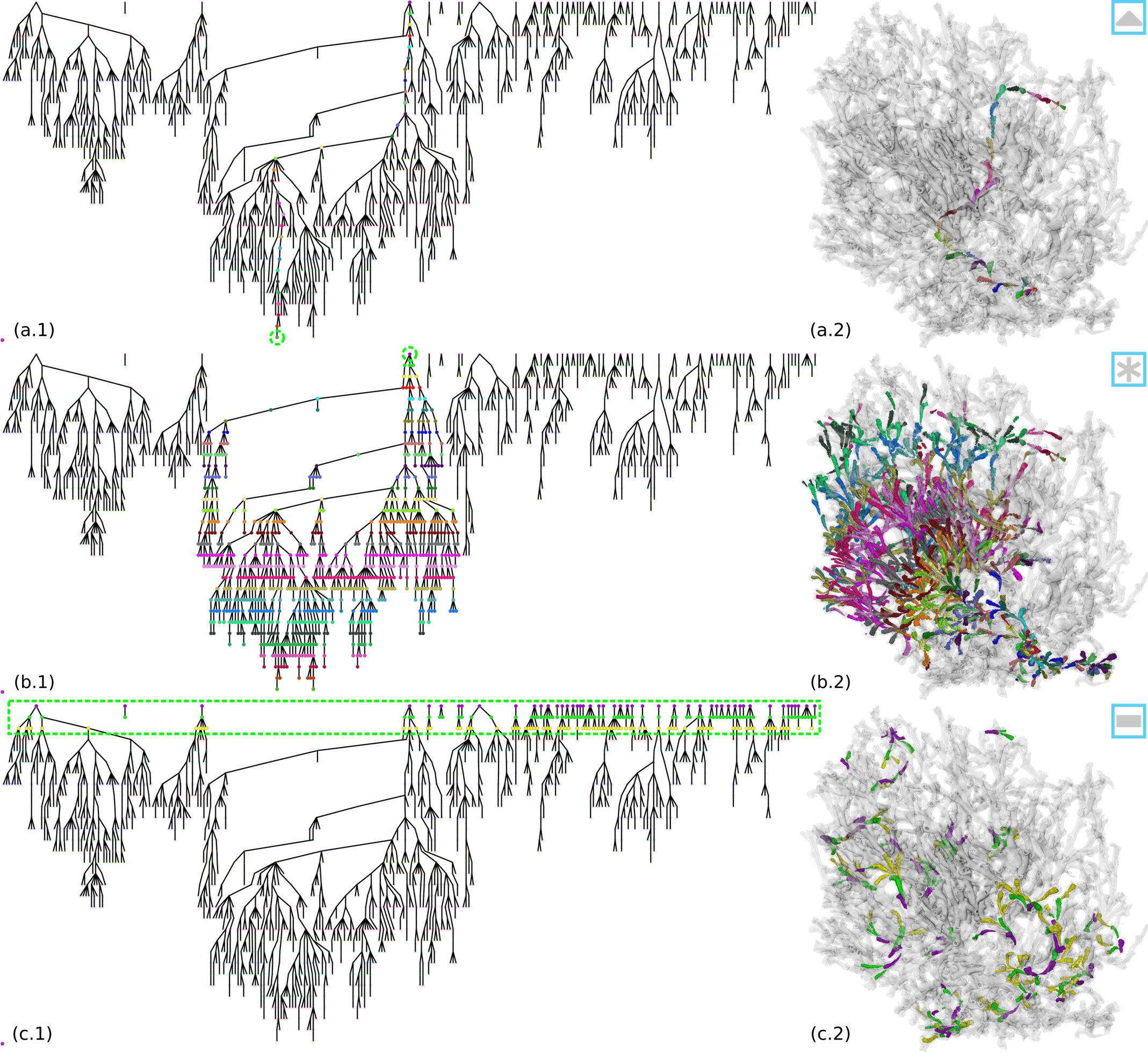}
    \caption{
        The images show different selections for the \textit{Lophelia pertusa} specimen \emph{SaM-ID43148}, similarly to the selections in \autoref{fig:cora selection tools}.
        \emph{(a)}%
            ~The dendrogram shows the selection of the longest ancestry path and the 3D visualization of the calices belonging to it.
        \emph{(b)}%
            ~The largest colony fragment is selected in the dendrogram (left) and visualized on the right side.
        \emph{(c)}%
            ~The three youngest generations are selected so that the 3D visualization gives an impression of (relative) location of the colony fragments present in the specimen.
    }
    \label{fig:sam-id43148 selections}
\end{figure*}

\end{document}